\documentclass{article} 
\usepackage[preprint]{neurips_2021}
\usepackage[utf8]{inputenc} 
\usepackage[T1]{fontenc}    
\usepackage{booktabs}       
\usepackage{amsfonts}       
\usepackage{nicefrac}       
\usepackage{microtype}      
\usepackage{xcolor}         
\usepackage{times}
\usepackage[colorlinks = true,
            linkcolor = blue,
            urlcolor  = blue,
            citecolor = blue,
            anchorcolor = blue]{hyperref}      
\usepackage{url}
\usepackage{latexsym}
\usepackage{amsmath}
\usepackage{graphicx} 
\usepackage{multirow}
\usepackage{multicol}
\usepackage{algorithm}    
\usepackage{algpseudocode}
\usepackage{amssymb}
\usepackage{ucs}
\usepackage{enumitem}
\usepackage{color}
\usepackage{subcaption}
\usepackage{amsmath}
\usepackage{amstext}
\usepackage{amsfonts}
\usepackage{bm}

\usepackage{geometry}

\usepackage{fancyhdr}
\usepackage{soul}

\usepackage{bbm}

\title{Lossless Acceleration for Seq2seq Generation with Aggressive Decoding}

\author{Tao Ge, ~~Heming Xia\thanks{Equal contribution: Xin Sun and Tao Ge contribute equally to Input-guided Aggressive Decoding; Heming Xia and Tao Ge contribute equally to Generalized Aggressive Decoding. This work was done during Xin Sun and Heming Xia's internship at MSR Asia. Correspondence to Tao Ge (\href{tage@microsoft.com}{tage@microsoft.com}).}, ~~Xin Sun$^*$, ~~Si-Qing Chen, ~~Furu Wei  \\
Microsoft\\
\texttt{\{tage, sqchen, fuwei\}@microsoft.com, \{xiaheming, sunx5\}@pku.edu.cn}
}

\begin{document}

\maketitle

\begin{abstract} 
We study lossless acceleration for seq2seq generation with a novel decoding algorithm -- Aggressive Decoding. Unlike the previous efforts (e.g., non-autoregressive decoding) speeding up seq2seq generation at the cost of quality loss, our approach aims to yield the identical (or better) generation compared with autoregressive decoding but in a significant speedup, achieved by innovative cooperation of aggressive decoding and verification that are both highly efficient due to parallel computing.

We propose two Aggressive Decoding paradigms for two kinds of seq2seq tasks: \textbf{1)} For the seq2seq tasks whose inputs and outputs are highly similar (e.g., Grammatical Error Correction), we propose Input-guided Aggressive Decoding that aggressively copies from the input sentence as drafted decoded tokens to verify in parallel; \textbf{2)} For other general seq2seq tasks (e.g., Machine Translation), we propose Generalized Aggressive Decoding that first employs an additional non-autoregressive decoding model for aggressive decoding and then verifies in parallel in the autoregressive manner.

We test Aggressive Decoding on the most popular 6-layer Transformer model on GPU in multiple seq2seq tasks: \textbf{1)} For Input-guided Aggressive Decoding, we show that it can introduce a $7\times$$\sim$$9\times$ speedup for the Transformer in Grammatical Error Correction and Text Simplification tasks with the identical results as greedy decoding; \textbf{2)} For Generalized Aggressive Decoding, we observe a $3\times$$\sim$$5\times$ speedup with the identical or even better quality in two important seq2seq tasks: Machine Translation and Abstractive Summarization. Moreover, Aggressive Decoding can benefit even more from stronger computing devices that are better at parallel computing. Given the lossless quality as well as significant and promising speedup, we believe Aggressive Decoding may potentially evolve into a \textit{de facto} standard for efficient and lossless seq2seq generation in the near future. Our codes are available at
\url{https://github.com/microsoft/unilm/tree/master/decoding}.
\end{abstract}

\section{Introduction}\label{sec:intro}
Despite many advantages of autoregressive decoding for seq2seq generation, it has been widely blamed for its inefficient inference, which sequentially decodes only one token at each step so that the next token prediction can condition on the previous decoding results.

To improve the inference efficiency for seq2seq generation, we propose Aggressive Decoding -- a novel efficient decoding paradigm. 
Different from the previous efforts (e.g., non-autoregressive decoding) speeding up seq2seq generation at the sacrifice of generation quality, Aggressive Decoding is a quality-lossless, aiming to yield the identical (or even better) generation compared with autoregressive decoding but in a significant speedup.

\begin{figure}[t]
    \centering
    \includegraphics[width=\textwidth]{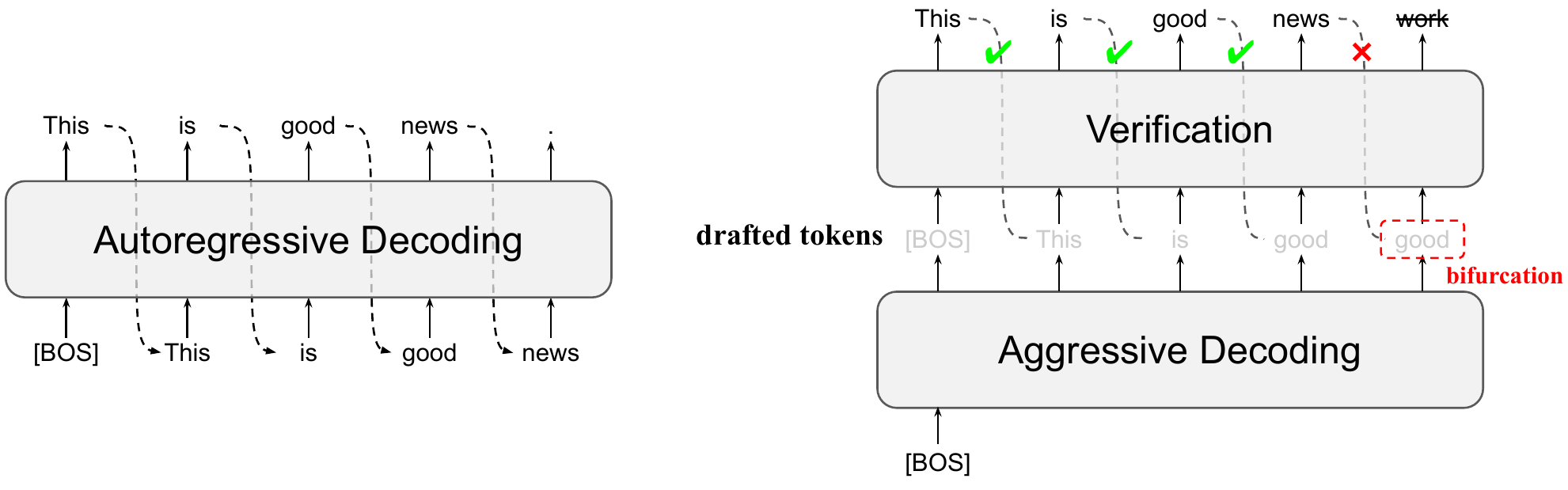}
    \caption{\textbf{Left}: Autoregressive Decoding that sequentially decodes only one token at each step so that the next token prediction can condition on the previous decoding results; \textbf{Right}: Aggressive Decoding that aggressively decodes a number of drafted tokens and verifies, which is not only highly efficient due to parallel computing, but also guarantees its generation is identical to autoregressive decoding.}
    \label{fig:ad_overview}
\end{figure}

The key idea of Aggressive Decoding is to replace the sequential decoding paradigm in autoregressive decoding with innovative cooperation of aggressive decoding and verification that can both efficiently perform in parallel: it first aggressively decodes as many tokens\footnote{We call them \textbf{drafted tokens} in this paper.} as possible; then it verifies the drafted tokens in the autoregressive manner where only the tokens that pass the verification can be accepted, as shown in Figure \ref{fig:ad_overview}. In this way, Aggressive Decoding can not only efficiently decode a number of tokens at each decoding iteration but also guarantee decoding results are identical to autoregressive decoding.

We propose two kinds of Aggressive Decoding approaches which differ in the way of drafted token decoding for different kinds of seq2seq tasks: \textbf{1)} For the seq2seq tasks whose inputs and outputs are highly similar (e.g., Grammatical Error Correction), we propose Input-guided Aggressive Decoding that aggressively uses the input sentence as drafted tokens to verify; \textbf{2)} For other seq2seq tasks that do not have highly similar inputs and outputs (e.g., Machine Translation), we propose Generalized Aggressive Decoding that first employs an additional non-autoregressive decoding model for generating the drafted tokens and then verifies.

We evaluate Aggressive Decoding on the most popular 6-layer Transformer model in multiple seq2seq tasks: \textbf{1)} For Input-guided Aggressive Decoding, we show that it can introduce a $7\times$$\sim$$9\times$ speedup in Grammatical Error Correction and Text Simplification tasks with identical results as greedy decoding; \textbf{2)} For Generalized Aggressive Decoding, we observe a $3\times$$\sim$$5\times$ speedup with the identical or even better quality as autoregressive decoding in Machine Translation and Abstractive Summarization. Moreover, we find Aggressive Decoding can benefit even more from more powerful computing devices that are better at parallel computing. Given the lossless quality as well as significant and promising speedup, we believe Aggressive Decoding may potentially evolve into a \textit{de facto} standard for efficient and lossless seq2seq generation in the near future, although it still has much room for improvement.

\section{Background: Autoregressive Scoring and Decoding in Transformer}\label{sec:background}

The Transformer is currently the most successful and widely used model architecture for seq2seq tasks such as Machine Translation, Abstractive Summarization and Grammatical Error Correction.

In the training phase, the Transformer learns an autoregressive scoring model $P(\boldsymbol{y}~|~\boldsymbol{x};\boldsymbol{\Phi})$, implemented with teacher forcing:
\begin{equation}\label{eq:train}
    \boldsymbol{\Phi^*} =\arg \max_{\boldsymbol{\Phi}} \log P(\boldsymbol{y}~|~\boldsymbol{x}; \boldsymbol{\Phi}) = \arg \max_{\boldsymbol{\Phi}} \sum_{i=0}^{l-1} \log P(y_{i+1}~|~\boldsymbol{y}_{\le i}, \boldsymbol{x}; \boldsymbol{\Phi})
\end{equation}
where $\boldsymbol{y}=(y_1, \dots, y_l)$ is the ground-truth target sequence and $\boldsymbol{y}_{\le i}=(y_0, \dots, y_i)$. As ground truth is available during training, Eq (\ref{eq:train}) can be efficiently obtained as the probability $ P(y_{i+1}~|~\boldsymbol{y}_{\le i}, \boldsymbol{x})$ at each step can be computed in parallel.

During inference, the output sequence $\boldsymbol{o}=(o_1, \dots, o_m)$ is derived by maximizing the following equation:

\begin{equation}
\boldsymbol{o^*} = \arg \max_{\boldsymbol{o}} \log P(\boldsymbol{o}~|~\boldsymbol{x}; \boldsymbol{\Phi}) = \arg \max_{\boldsymbol{o}} \sum_{j=0}^{m-1} \log P(o_{j+1}~|~\boldsymbol{o}_{\le j}, \boldsymbol{x}; \boldsymbol{\Phi})
\end{equation}

However, since no ground truth is available in the inference phase, the model has to autoregressively decode only one token at each iteration conditioning on the previous decoded tokens $\boldsymbol{o}_{\le j}$ instead of predicting in parallel as in the training phase. As a result, autoregressive decoding needs to sequentially iterate $m$ steps to decode a sequence of length $m$, which largely limits computational parallelism, becoming the main bottleneck of inference efficiency. 

\section{Aggressive Decoding}

As Section \ref{sec:intro} introduces, Aggressive Decoding aims to accelerate autoregressive (AR) decoding without quality loss, achieved by cooperation of aggressive decoding and verification, which may work in a different way for different kinds of seq2seq tasks.

Specifically, we start with Input-guided Aggressive Decoding (IAD) in Section \ref{subsec:input-guided} that is straightforward for the seq2seq tasks whose inputs and outputs are highly similar. Then, we generalize Aggressive Decoding for other seq2seq tasks by proposing Generalized Aggressive Decoding (GAD) in Section \ref{subsec:gad}.

\subsection{Input-guided Aggressive Decoding}\label{subsec:input-guided}

As introduced in Section \ref{sec:background}, the Transformer decodes only one token at each step during inference. For the seq2seq tasks whose output sequence is usually very similar to the input, fortunately, we can simply assume the model's outputs are identical to the inputs, which allows us to aggressively decode a number of drafted tokens by directly copying from the input, motivating us to propose Input-guided Aggressive Decoding. 

\begin{figure}[t]
    \centering
    \includegraphics[width=\textwidth]{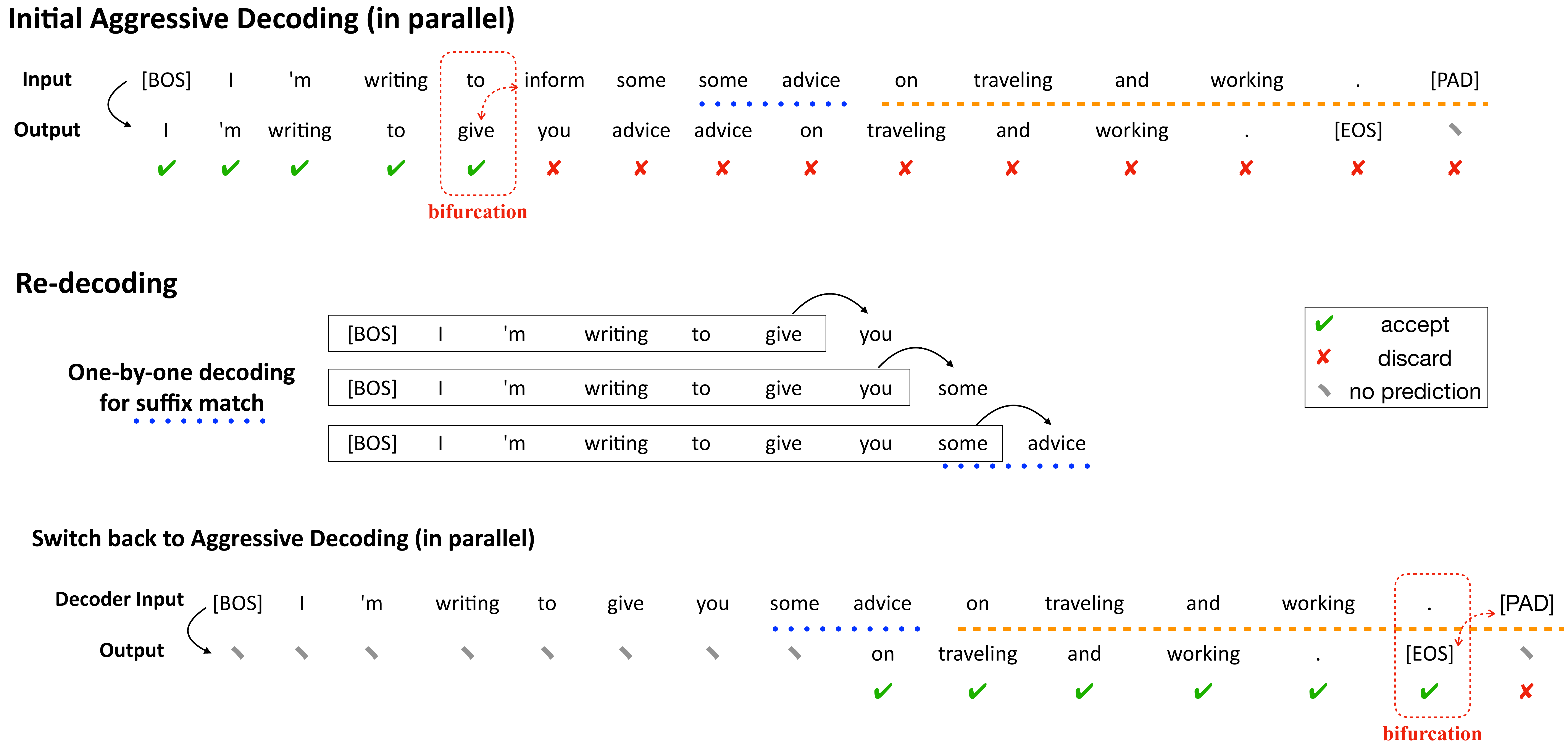} 
    \caption{The overview of Input-guided Aggressive Decoding. Input-guided Aggressive Decoding tries decoding as many tokens as possible in parallel with the assumption that the input and output should be identical. If we find a bifurcation when we verify the drafted tokens, then we accept the predictions before (including) the bifurcation, and discard all the predictions after the bifurcation and re-decode them using original one-by-one autoregressive decoding. If we find a suffix match (i.e., \textit{some advice} highlighted with the \textcolor{blue}{blue dot lines}) between the output and the input during one-by-one re-decoding, we switch back to Input-guided Aggressive Decoding by copying the tokens (highlighted with the \textcolor{orange}{orange dashed lines}) following the matched tokens in the input to the decoder input by assuming they will be the same.}
    \label{fig:ig_ag}
\end{figure}

The overview of Input-guided Aggressive Decoding is shown in Figure \ref{fig:ig_ag}, which involves initial aggressive decoding and re-decoding. We use Grammatical Error Correction (GEC) as an example task to demonstrate how Input-guided Aggressive Decoding works in the following subsections.

\subsubsection{Initial Aggressive Decoding} \label{subsubsec:iag}
The key motivation of Input-guided Aggressive Decoding is the assumption that the output sequence $\boldsymbol{o}=(o_1, \dots, o_m)$ should be almost the same with the input sequence $\boldsymbol{x}=(x_1, \dots, x_n)$. At the initial step, instead of only decoding the first token $o_1$ conditioning on the special [\textit{BOS}] token $o_0$, Input-guided Aggressive Decoding decodes $\boldsymbol{o}_{1 \dots n}$ conditioning on the drafted tokens $\boldsymbol{\hat{o}}_{0 \dots n-1}$ in parallel with the assumption that $\boldsymbol{\hat{o}}_{0 \dots n-1} = \boldsymbol{x}_{0, \dots, n-1}$. Specifically, for $j \in \{0, 1, \dots, n-2, n-1\}$, $o_{j+1}$ is decoded as follows:

\begin{equation}
\begin{split}
o_{j+1}^* &= \arg \max_{o_{j+1}} \log P(o_{j+1}~|\boldsymbol{o}_{\le j}, \boldsymbol{x}; \boldsymbol{\Phi}) \\ & = \arg \max_{o_{j+1}} \log P(o_{j+1}~|~\boldsymbol{\hat{o}}_{\le j}, \boldsymbol{x}; \boldsymbol{\Phi}) \\ 
& = \arg \max_{o_{j+1}} \log P(o_{j+1}~|~\boldsymbol{x}_{\le j}, \boldsymbol{x}; \boldsymbol{\Phi})
\end{split}
\end{equation}
where $\boldsymbol{\hat{o}}_{\le j}$ is the previous drafted tokens for step $j+1$, which is assumed to be the same with $\boldsymbol{x}_{\le j}$.

After we obtain $\boldsymbol{o}_{1...n}$, we verify whether $\boldsymbol{x}_{1...n}$ is actually identical to $\boldsymbol{o}_{1...n}$ or not. If $\boldsymbol{x}_{1...n}$ is fortunately exactly the same as $\boldsymbol{o}_{1...n}$, the inference will finish, meaning that the model finds no grammatical errors in the input sequence $\boldsymbol{x}_{1...n}$ and keeps the input untouched. In more cases, however, $\boldsymbol{x}_{1...n}$ will not be exactly the same as $\boldsymbol{o}_{1...n}$. In such a case, we have to stop to find the first bifurcation position $k$ so that $\boldsymbol{o}_{1...k-1} = \boldsymbol{x}_{1...k-1}$ and $o_k \neq x_k$.

Since $\boldsymbol{o}_{1...k-1} = \boldsymbol{\hat{o}}_{1...k-1} = \boldsymbol{x}_{1...k-1}$, the predictions $\boldsymbol{o}_{1...k}$ could be accepted as they will not be different even if they are decoded through the original autoregressive greedy decoding. However, for the predictions $\boldsymbol{o}_{k+1 \dots n}$, we have to discard and re-decode them because $o_k \neq \hat{o}_k$.

\subsubsection{Re-decoding} \label{subsubsec:rdra}  
As $o_k \neq \hat{o}_k = x_k$, we have to re-decode for $o_{j+1}$ ($j \ge k$) one by one following the original autoregressive decoding:

\begin{equation}
o_{j+1}^*= \arg \max_{o_{j+1}} P(o_{j+1}~|~\boldsymbol{o}_{\le j}, \boldsymbol{x}; \boldsymbol{\Phi})
\end{equation}

After we obtain $\boldsymbol{o}_{\le j}$ ($j>k$), we try to match its suffix to the input sequence $\boldsymbol{x}$ for further aggressive decoding. If we find its suffix $\boldsymbol{o}_{j-q \dots j}$ ($q \ge 0$) is the unique substring of $\boldsymbol{x}$ such that $\boldsymbol{o}_{j-q \dots j}=\boldsymbol{x}_{i-q \dots i}$, then we can assume that $\boldsymbol{o}_{j+1 \dots}$ will be very likely to be the same with $\boldsymbol{x}_{i+1 \dots}$ because of the special characteristic of the task that the input and output are highly similar.

If we fortunately find such a suffix match, then we can switch back to Input-guided Aggressive Decoding to decode in parallel with the assumption $\boldsymbol{\hat{o}}_{j+1 \dots} = \boldsymbol{x}_{i+1 \dots}$. Specifically, the token $o_{j+t}$ ($t>0$) is decoded as follows:

\begin{equation} \label{eq:agcore}
    o_{j+t}^* = \arg \max_{o_{j+t}} P(o_{j+t}~|~\boldsymbol{o}_{< j+t}, \boldsymbol{x}; \boldsymbol{\Phi}) 
\end{equation}
In Eq (\ref{eq:agcore}), $\boldsymbol{o}_{<j+t}$ is derived as follows:
\begin{equation} \label{eq:agcore1}
\boldsymbol{o}_{<j+t} = \textsc{Cat}(\boldsymbol{o}_{\le j}, \boldsymbol{\hat{o}}_{j+1 \dots j+t-1}) = \textsc{Cat}(\boldsymbol{o}_{\le j}, \boldsymbol{x}_{i+1 \dots i+t-1}) 
\end{equation}
where \textsc{Cat}($\boldsymbol{a}$, $\boldsymbol{b}$) is the operation that concatenates two sequences $\boldsymbol{a}$ and $\boldsymbol{b}$.

Otherwise (i.e., we cannot find a suffix match at the step), we continue decoding using the original autoregressive greedy decoding approach until we find a suffix match.

We summarize the process of Input-guided Aggressive Decoding in Algorithm \ref{alg:ag}. For simplifying implementation, we make minor changes in Algorithm \ref{alg:ag}: 1) we set $o_0=x_0=\textrm{[}BOS\textrm{]}$ in Algorithm \ref{alg:ag}, which enables us to regard the initial aggressive decoding as the result of suffix match of $o_0=x_0$; 2) we append a special token $\textrm{[}PAD\textrm{]}$ to the end of $\boldsymbol{x}$ so that the bifurcation (in the $5^{th}$ line in Algorithm \ref{alg:ag}) must exist (see the bottom example in Figure \ref{fig:ig_ag}). Since we discard all the computations and predictions after the bifurcation for re-decoding, Input-guided Aggressive Decoding guarantees that generation results are exactly the same as greedy decoding (i.e., beam=1). However, as Input-guided Aggressive Decoding decodes many tokens in parallel, it largely improves the computational parallelism during inference, greatly benefiting the inference efficiency.

\begin{algorithm}[t]
\caption{Input-guided Aggressive Decoding\label{alg:ag}}
\textbf{Input:} $\boldsymbol{\Phi}$, $\boldsymbol{x}=(\textrm{[}BOS\textrm{]}, x_1, \dots, x_n, \textrm{[}PAD\textrm{]})$, $\boldsymbol{o}=(o_0)=(\textrm{[}BOS\textrm{]})$; \\
\textbf{Output:} $\boldsymbol{o}_{1 \dots j}=(o_1, \dots, o_j)$;
\begin{algorithmic}[1]
\State Initialize $j \gets 0$;
\While{$o_j \neq$ [$EOS$] and $j <$ MAX\_LEN}
\If{$\boldsymbol{o}_{j-q \dots j}$ $(q \ge 0)$ is a unique substring of $\boldsymbol{x}$ such that $\exists~!~i:   \boldsymbol{o}_{j-q \dots j} = \boldsymbol{x}_{i-q \dots i}$}
\State{Aggressive Decode $\boldsymbol{\widetilde{o}}_{j+1 \dots}$ according to Eq (\ref{eq:agcore}) and Eq (\ref{eq:agcore1})};
\State Find bifurcation $j+k$ ($k>0$) such that $\boldsymbol{\widetilde{o}}_{j+1 \dots j+k-1} = \boldsymbol{x}_{i+1 \dots i+k-1}$ and ${\widetilde{o}}_{j+k} \neq x_{i+k}$;
\State $\boldsymbol{o} \gets \textsc{Cat}(\boldsymbol{o}, \boldsymbol{\widetilde{o}}_{j+1 \dots j+k})$;
\State $j \gets j+k$;
\Else 
\State Decode $o_{j+1}^* = \arg \max_{o_{j+1}} P(o_{j+1}~|~\boldsymbol{o}_{\le j}, \boldsymbol{x}; \boldsymbol{\Phi})$;
\State $\boldsymbol{o} \gets \textsc{Cat}(\boldsymbol{o}, o_{j+1}^*)$;
\State $j \gets j+1$;
\EndIf
\EndWhile
\end{algorithmic}
\end{algorithm}

\subsection{Generalized Aggressive Decoding}\label{subsec:gad}
Although Input-guided Aggressive Decoding is intuitive for the seq2seq tasks like GEC whose inputs and outputs are similar, it cannot be applied to other seq2seq tasks that are not characterized by highly similar inputs and outputs. To generalize Aggressive Decoding for general seq2seq tasks like Machine Translation (MT), we propose Generalized Aggressive Decoding (GAD).

\begin{figure}[t]
\centering
\includegraphics[width=1.0\textwidth]{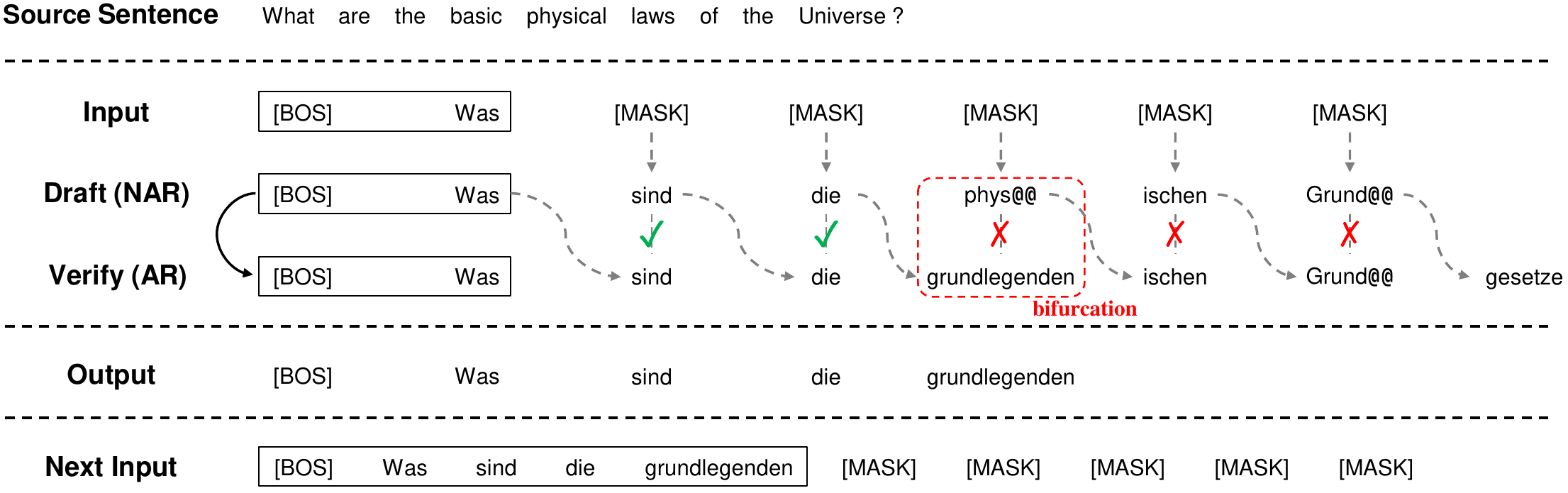}
\caption{Generalized Aggressive Decoding where a decoding iteration is involved with two substeps: \textit{draft} and \textit{verify}. In the \textbf{Draft} substep, an NAR model drafts (i.e., decodes) a block (block size $k=5$ for this example) of tokens in parallel conditioning on the source sentence and previous decoded tokens (i.e., the tokens in the rectangle boxes). In the \textbf{Verify} substep, drafted tokens are verified in parallel: bifurcation is detected as the first position where we find the drafted token does not match the top-1 result verified by an AR model. The drafted tokens after the bifurcation position are all discarded, for guaranteeing GAD's decoding results will be the exactly same with (AR) greedy decoding.}
\label{fig:GAD}
\end{figure}

Unlike Input-guided Aggressive Decoding which simply copies from the input sentence as the drafted tokens, GAD additionally employs a non-autoregressive (NAR) decoding model to aggressively generate the drafted tokens efficiently in parallel, allowing GAD to be applicable to any seq2seq task where NAR works well. Specifically, GAD decomposes every decoding iteration into two substeps -- \textit{draft} and \textit{verify}:

\paragraph{Draft} At each decoding iteration, GAD first utilizes an NAR model to aggressively decode a number of drafted tokens (denoted as \texttt{[MASK]} in its decoder input in Figure \ref{fig:GAD}) in parallel, conditioning on preceding decoded tokens. Formally, given the source sentence $\bm{x}={\left(x_1, x_2, \ldots, x_n\right)}$ and the previous decoded tokens $\bm{o}_{\le j}={\left(o_1, o_2, \ldots, o_j\right)}$, GAD decodes the next $k$ (drafted) tokens as a block in parallel:

\begin{equation}\label{eq:draft-infer}
\nonumber
\widetilde{\bm{o}}_{j+1 \cdots j+k} = \arg \max _{\widetilde{\bm{o}}_{j+1 \cdots j+k}} \sum_{i=1}^{k} \log P\left(\widetilde{o}_{j+i} \mid \bm{o}_{\le j}, \bm{x};\bm{\Phi}_{\textrm{NAR}}\right)
\end{equation}

\paragraph{Verify} Then, the drafted tokens $\widetilde{\bm{o}}_{j+1 \cdots j+k}$ are verified with an AR model in the autoregressive manner, which performs in parallel. As Input-guided Aggressive Decoding, we find the bifurcation position $c$ by comparing the drafted tokens with the AR prediction results conditioning on the draft as Figure \ref{fig:GAD} shows:

\begin{equation}
\nonumber
c = \arg \max_{i} \frac{\mathbbm{1}\left(\widetilde{o}_{j+i} \neq \hat{o}_{j+i}\right)}{i}, 1 \le i \le k
\end{equation}

\begin{equation}\label{eq:verifier}
    \hat{o}_{j+i} = \arg \max_{\hat{o}_{j+i}} \log P(\hat{o}_{j+i}|\bm{o_{\le j}}, \widetilde{\bm{o}}_{j+1 \cdots j+i-1}, \bm{x}; \bm{\Phi}_{\textrm{AR}})
\end{equation}
where $\mathbbm{1}(\cdot)$ is the indicator function and $\hat{o}_{j+i}$ is the top-1 result verified by the AR model conditioning on the previous decoded tokens $\bm{o}_{\le j}$ and the drafted tokens $\widetilde{\bm{o}}_{j+1 \cdots j+i-1}$. We only accept the verified tokens before (including) the bifurcation position as decoded tokens, which ensures GAD to yield the same results as greedy decoding of AR:

\begin{equation}
\nonumber
    \bm{o}_{j+1 \cdots j+c} = \bm{\hat{o}}_{j+1 \cdots j+c} = (\bm{\widetilde{o}}_{j+1 \cdots j+c-1}, \hat{o}_{j+c})
\end{equation}

We iterate decoding with the above substeps until the termination condition is met, i.e. the \texttt{[EOS]} token is decoded or the sentence reaches the maximal length. As illustrated, GAD is highly efficient because both \textit{draft} and \textit{verify} perform in parallel. 

\subsubsection{NAR drafter}\label{subsec:nat}
As demonstrated above, an NAR model is the key to generalizing Aggressive Decoding to general seq2seq tasks, which can efficiently generate drafted tokens in parallel. Our NAR drafter differs from other NAR models in two aspects: First, we only require the NAR drafter to decode a block (i.e., fixed length) of tokens in each decoding iteration, instead of the whole sequence; Second, as illustrated in Figure \ref{fig:GAD}, since we decode from \textit{Left} to \textit{Right}, the NAR drafter is required to decode tokens conditioning on the previously decoded tokens. Formally, given the source sentence $\bm{x}=(x_1,\cdots,x_n)$ and the randomly sampled prefix $\bm{y}_{0 \cdots p}$ ($0 \le p < m$)  of the target sentence $\bm{y}=(y_1,\cdots,y_m)$, the model is trained to predict the next $k$ tokens, as shown in Figure \ref{fig:GAD}:

\begin{equation}
\nonumber
\mathcal{L}_{\textrm{NAR}}=\sum_{i=p+1}^{p+k} \log P\left(y_{i} | \bm{y}_{1 \cdots p}, \bm{x}; \bm{\Phi}_{\textrm{NAR}}\right)
\end{equation}

In addition, we leverage the glancing strategy following \cite{Qian:2020}, which exploits curriculum learning during training to get better performance. As in previous NAR work, we apply sequence-level knowledge distillation (Seq-KD)~\citep{kim2016sequence} by an AR Transformer teacher model to train our NAR drafter.

\subsubsection{AR verifier}
\label{sec:AT-verifier}
We use the conventional autoregressive Transformer (see Section \ref{sec:background}) as our AR verifier, which is the key to guaranteeing the generation quality. As we hope as many drafted tokens by the NAR model as possible can be accepted by the AR verifier for a higher speedup, we also apply Seq-KD to the AR verifier by a shared teacher (with the NAR drafter), which not only allows the NAR drafter and AR verifier to perform similarly, but also helps improve the AR verifier's prediction quality~\citep{furlanello2018born}.

\subsubsection{GAD++}
\label{sec:GAD++}
As shown in Figure \ref{fig:GAD} and discussed above, the vanilla GAD only accepts the drafted tokens that match the top-1 result of the AR verifier, which guarantees that GAD's generation is identical to greedy decoding of AR. However, the top-1 results are not necessarily better than the drafted tokens. As a result, the strict verification criterion (i.e., top-1 matching) will result in many good drafted tokens being discarded just because they are different from the top-1 result of the AR verifier, which limits the speedup of GAD.

\begin{figure}[t]
\centering
\includegraphics[width=1.0\textwidth]{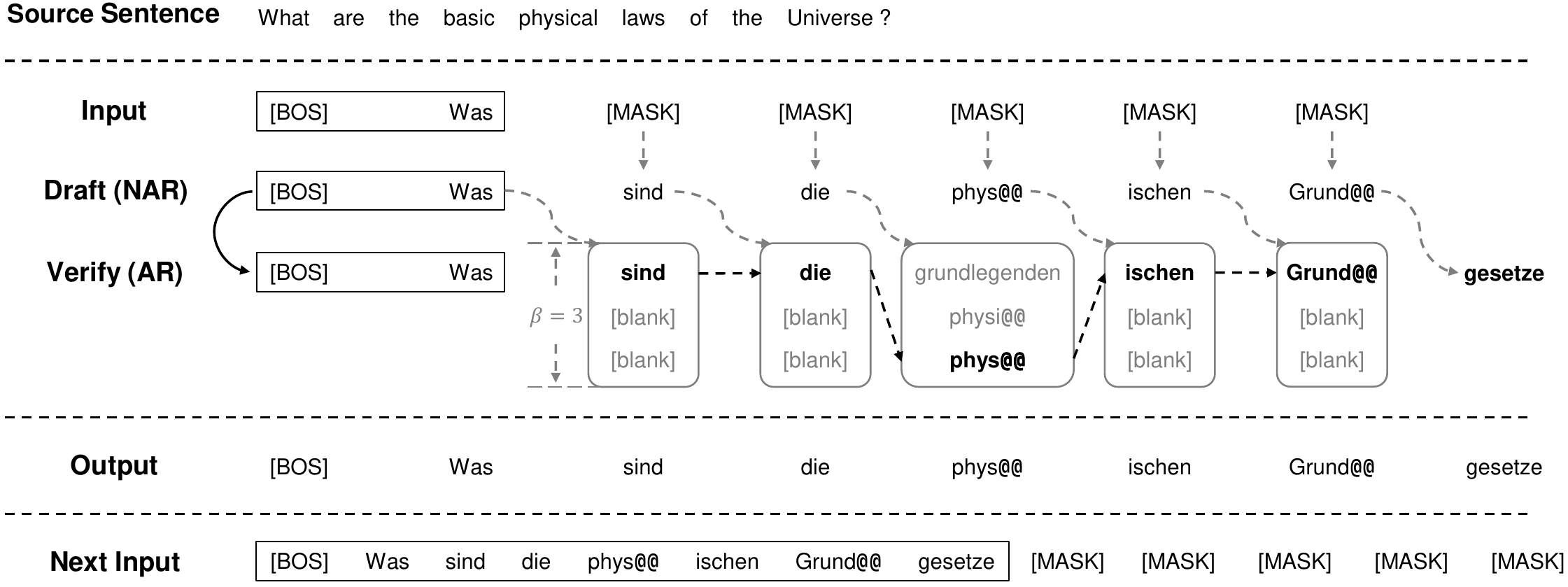}
\caption{Illustration of GAD++. Compared to the vanilla GAD strictly requiring the drafted tokens to match the top-1 result of the AR verifier, GAD++ slightly loosens the criterion to trust NAR's draft more, by only requiring the drafted tokens to fall in the \textit{top}-$\beta$ of the AR verifier with a tolerable log-likelihood gap (not shown in this Figure; see Eq (\ref{eq:gad++2})). As a result, GAD++ allows more drafted tokens to be accepted even if they are slightly different from the top-1 result of the AR verifier, leading to a higher inference speedup.}
\label{fig:GAD++}
\end{figure}

To overcome this limitation, we propose a variant of GAD named GAD++, which is illustrated in Figure \ref{fig:GAD++}. Instead of the rigid top-1 matching requirement in the vanilla GAD shown in Eq (\ref{eq:verifier}) , GAD++ loosens the criterion to trust the NAR's draft more, by only requiring the drafted tokens to fall in \textit{top-$\beta$} candidates with a tolerable (log-likelihood) score gap $\tau$ (away from the top-1 result):

\begin{equation}
\nonumber
    \hat{o}_{j+i}=
    \begin{cases}
        \widetilde{o}_{j+i}, ~\textrm{if}~ \textit{GAD++ requirement}~ \textrm{is met}  \\
        \textrm{same as Eq (\ref{eq:verifier})}, ~\textrm{otherwise}
    \end{cases}
\end{equation}

As discussed above, \textit{GAD++ requirement} is met if Eq (\ref{eq:gad++1}) and (\ref{eq:gad++2}) are both true:
\begin{equation}\label{eq:gad++1}
   \log P(\widetilde{o}_{j+i}|\cdot) \ge \log P(\hat{o}_{j+i}^{(\beta)}|\cdot)
\end{equation}
\begin{equation}\label{eq:gad++2}
    \log P(\hat{o}_{j+i}^{(1)}|\cdot) - \log P(\widetilde{o}_{j+i}|\cdot) \le \tau
\end{equation}
where $\log P(\hat{o}_{j+i}^{(\beta)}|\cdot)$ is the top-$\beta$ ranked result's log-likelihood score by the AR verifier.

The advanced verification criterion with the hyperparameter top-$\beta$ and tolerance $\tau$ not only allows more drafted tokens to be accepted for a higher speedup but also enables GAD++ to generate beyond greedy decoding.

\section{Experiments}\label{sec:exp}
We test Aggressive Decoding in various seq2seq tasks: For Input-guided Aggressive Decoding, we mainly evaluate in Grammatical Error Correction (Section \ref{subsec:iad_exp}); while for Generalized Aggressive Decoding, we choose Machine Translation as the main task for evaluation (Section \ref{subsec:gad_exp}).

\subsection{Evaluations for Input-guided Aggressive Decoding}\label{subsec:iad_exp}
\subsubsection{Data and Model Configuration}
We follow recent work in English GEC to conduct experiments in the restricted training setting of BEA-2019 GEC shared task~\citep{bryant2019bea}: We use Lang-8 Corpus of Learner English~\citep{mizumoto2011mining}, NUCLE~\citep{dahlmeier2013building}, FCE~\citep{yannakoudakis2011new} and W\&I+LOCNESS~\citep{granger1998computer, bryant2019bea} as our GEC training data. 
For facilitating fair comparison in the efficiency evaluation, we follow the previous studies~\citep{omelianchuk2020gector,chen2020improving} which conduct GEC efficiency evaluation to use CoNLL-2014~\citep{ng2014conll} dataset that contains 1,312 sentences as our main test set, and evaluate the speedup as well as Max-Match~\citep{dahlmeier2012better} precision, recall and $F_{0.5}$ using their official evaluation scripts\footnote{\url{https://github.com/nusnlp/m2scorer}}. For validation, we use CoNLL-2013~\citep{ng-etal-2013-conll} that contains 1,381 sentences as our validation set.
To compare with the state-of-the-art approaches in English GEC that pretrain with synthetic data, we also synthesize 300M error-corrected sentence pairs for pretraining the English GEC model following the approaches of \cite{grundkiewicz2019neural} and \cite{zhang2019sequence}. Note that in the following evaluation sections, the models evaluated are by default trained without the synthetic data unless they are explicitly mentioned.   

We use the most popular GEC model architecture -- Transformer-big~\citep{vaswani2017attention} as our baseline model which has a 6-layer encoder and 6-layer decoder with 1,024/4,096 embedding/FFN dimension. We train the English GEC model using an encoder-decoder shared vocabulary of 32K Byte Pair Encoding~\citep{sennrich2016neural} tokens. We include more training details in the supplementary notes. 

All the efficiency evaluations are conducted in the online inference setting (i.e., batch size=1) by default. We perform model inference with fairseq\footnote{\url{https://github.com/pytorch/fairseq}} implementation using Pytorch 1.5.1 with 1 Nvidia Tesla V100-PCIe of 16GB GPU memory under CUDA 10.2.

\begin{table}[t] 
\centering
\small
\begin{tabular}{l|c|c|c|ccc}
\hline
\multirow{2}{*}{\bf Model}       & \multirow{2}{*}{\bf Synthetic Data} & \multirow{2}{*}{\bf Total Latency (s)} & \multirow{2}{*}{\bf Speedup} & \multicolumn{3}{c}{ \textbf{CoNLL-13}} \\  
 & & &  & $P$ & $R$ & $F_{0.5}$ \\ \hline
Transformer-big (beam=5) & No & 440 & 1.0$\times$ & \bf 53.84 & 18.00 & \bf 38.50 \\
Transformer-big (beam=1) & No & 397 & 1.1$\times$ & 52.75 & \bf 18.34 & 38.36 \\
Transformer-big (IAD) & No & \bf 54 & \bf \boldmath 8.1$\times$ & 52.75 & \bf 18.34 & 38.36 \\ \hline
Transformer-big (beam=5) & Yes & 437 & 1.0$\times$ & \bf 57.06 & 23.62 &  44.47 \\
Transformer-big (beam=1) & Yes & 390 & 1.1$\times$ & 56.45 & \bf 24.70 & \bf 44.91     \\
Transformer-big (IAD) & Yes & \bf 60 & \bf \boldmath 7.3$\times$ & 56.45 & \bf 24.70 & \bf 44.91     \\ \hline
\end{tabular}
\vspace{0.2cm}
\caption{The performance and online inference efficiency of the Transformer-big with Input-guided Aggressive Decoding (IAD) in our validation set (CoNLL-13) that contains 1,381 sentences. We use Transformer-big (beam=5) as the baseline to compare the performance and efficiency of IAD. \label{tab:agresult1}}
\end{table}

\subsubsection{Results}

We evaluate Input-guided Aggressive Decoding (IAD) in our validation set (CoNLL-13) which contains 1,381 validation examples. As shown in Table \ref{tab:agresult1}, IAD achieves an over $7\times$ speedup over the autoregressive beam search (beam=5) baseline, and generates exactly the same predictions as greedy decoding, as discussed in Section \ref{subsec:input-guided}. Since greedy decoding can achieve comparable overall performance (i.e., $F_{0.5}$) with beam search while it tends to make more edits resulting in higher recall but lower precision, the advantage of IAD in practical GEC applications is obvious given its strong performance and superior efficiency.

\begin{figure}[t]
    \centering
    \includegraphics[width=10cm]{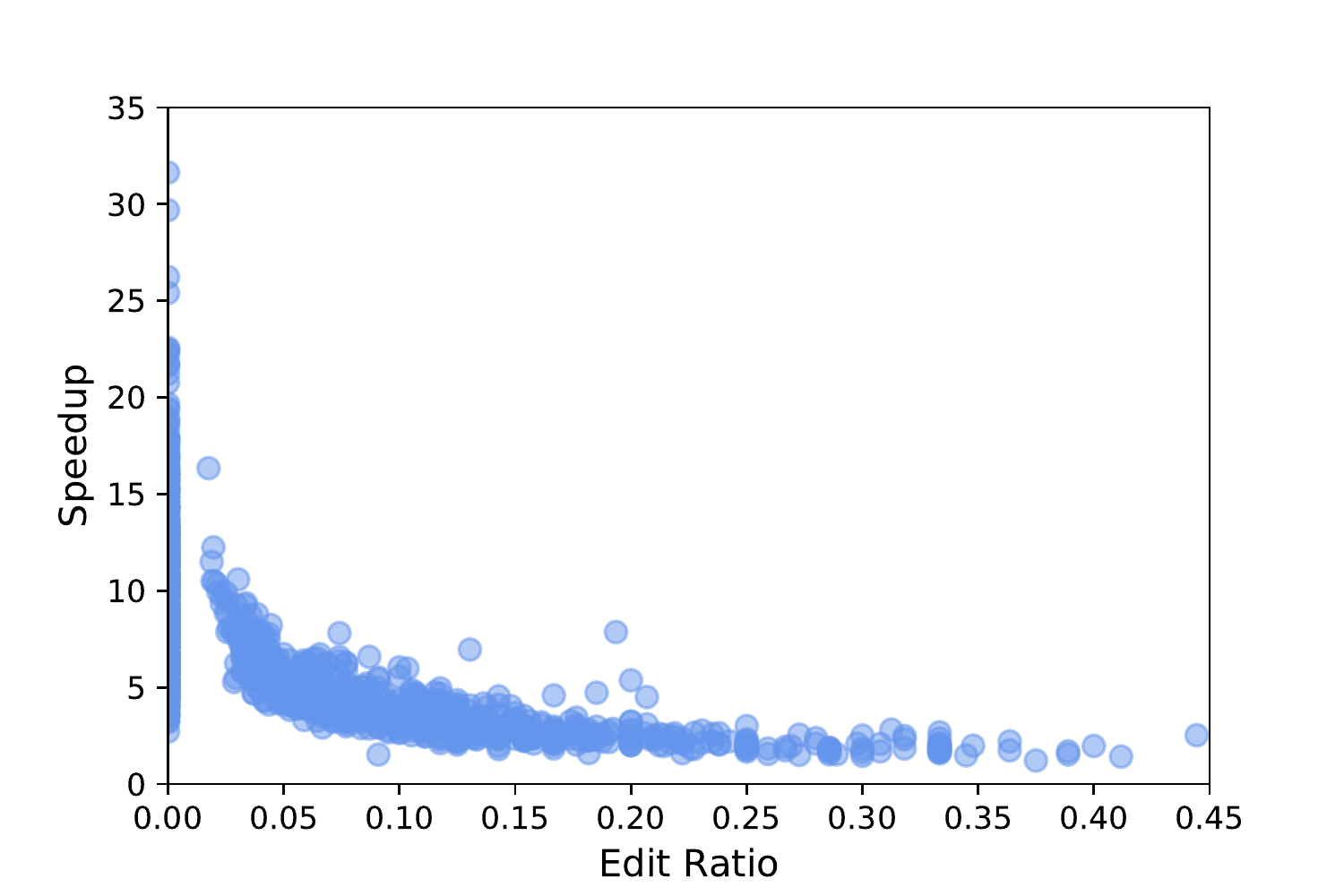}
    \caption{The speedup (over greedy decoding) distribution of all the 1,381 validation examples with respect to their edit ratio in CoNLL-13.}
    \label{fig:density}
\end{figure}

\begin{table}[t] 
\centering
\small
\scalebox{0.96}{
\begin{tabular}{c|c|p{5.1cm}|p{5.2cm}}
\hline
\bf Speedup & \bf Edit Ratio & \bf Input & \bf Output \\
 \hline
16.7$\times$ & 0 & Personally , I think surveillance technology such as RFID ( radio-frequency identification ) should not be used to track people , for the benefit it brings to me can not match the concerns it causes . & [\textcolor{blue}{Personally , I think surveillance technology such as RFID ( radio-frequency identification ) should not be used to track people , for the benefit it brings to me can not match the concerns it causes .}]$_0$ \\
 \hline
 5.8$\times$ & 0 & Nowadays , people use the all-purpose smart phone for communicating . & [\textcolor{blue}{Nowadays , people use the all-purpose smart phone for communicating .}]$_0$ \\ 
 \hline
 6.8$\times$ & 0.03 & Because that the birth rate is reduced while the death rate is also reduced , the percentage of the elderly is increased while that of the youth is decreased . & [\textcolor{blue}{Because the}]$_0$ [\textcolor{red}{birth}]$_1$ [\textcolor{blue}{rate is reduced while the death rate is also reduced , the percentage of the elderly is increased while that of the youth is decreased .}]$_2$ \\
 \hline
 5.1$\times$ & 0.06 & More importantly , they can share their ideas of how to keep healthy through Internet , to make more interested people get involve and find ways to make life longer and more wonderful . & [\textcolor{blue}{More importantly , they can share their ideas of how to keep healthy through the}]$_0$ [\textcolor{red}{Internet}]$_1$ [\textcolor{blue}{, to make more interested people get involved}]$_2$ [\textcolor{red}{and}]$_3$ [\textcolor{red}{find}]$_4$ [\textcolor{blue}{ways to make life longer and more wonderful .}]$_5$   \\
 \hline
 3.5$\times$ & 0.13 & As a result , people have more time to enjoy advantage of modern life . &  [\textcolor{blue}{As a result , people have more time to enjoy the}]$_0$ [\textcolor{red}{advantages}]$_1$ [\textcolor{red}{of}]$_2$ [\textcolor{blue}{modern life .}]$_3$  \\
 \hline
 1.5$\times$ & 0.27 & Nowadays , technology is more advance than the past time . & [\textcolor{blue}{Nowadays , technology is more advanced}]$_0$ [\textcolor{red}{than}]$_1$ [\textcolor{blue}{in}]$_2$ [\textcolor{red}{the}]$_3$ [\textcolor{blue}{past .}]$_4$ \\
 \hline
 1.4$\times$ & 0.41 & People are able to predicate some disasters like the earth quake and do the prevention beforehand . & [\textcolor{blue}{People are able to predict}]$_0$ [\textcolor{red}{disasters}]$_1$ [\textcolor{blue}{like the earthquake}]$_2$ [\textcolor{red}{and}]$_3$ [\textcolor{blue}{prevent}]$_4$ [\textcolor{red}{them}]$_5$ [\textcolor{red}{beforehand}]$_6$ [\textcolor{blue}{.}]$_7$\\
  \hline
\end{tabular}
}
\vspace{0.1cm}
\caption{Examples of various speedup ratios by IAD over greedy decoding in CoNLL-13. We show how the examples are decoded in the column of \textbf{Output}, where the tokens within a \textcolor{blue}{blue block} are decoded in parallel through aggressive decoding while the tokens in \textcolor{red}{red blocks} are decoded through the original autoregressive greedy decoding.\label{tab:casestudy}}
\end{table}

We further look into the efficiency improvement by IAD. Figure \ref{fig:density} shows the speedup distribution of the 1,381 examples in CoNLL-13 with respect to their edit ratio which is defined as the normalized (by the input length) edit distance between the input and output. It is obvious that the sentences with fewer edits tend to achieve higher speedup, which is consistent with our intuition that most tokens in such sentences can be decoded in parallel through IAD; on the other hand, for the sentences that are heavily edited, their speedup is limited because of frequent re-decoding. To give a more intuitive analysis, we also present concrete examples with various speedup in our validation set to understand how IAD improves the inference efficiency in Table \ref{tab:casestudy}.

\begin{table}[t]
\centering
\small
\begin{tabular}{l|c|c|cc}
\hline
\multirow{2}{*}{\textbf{Model}} & \multirow{2}{*}{\textbf{Synthetic Data}} & \bf Multi-stage & \multicolumn{2}{c}{\textbf{CoNLL-14}} \\ 
 & & \bf Fine-tuning  & $F_{0.5}$ & Speedup \\ 
 \hline
\it Transformer-big (beam=5) & Yes & No & 61.6 & 1.0$\times$ \\ \hline
\textit{PIE}$^\star$~\citep{awasthi2019parallel} & Yes & No  & 59.7 & \underline{10.3$\times$} \\
\textit{Span Correction}$^\star$~\citep{chen2020improving}  & Yes & No & 61.0 & 2.6$\times$ \\
\textit{Seq2Edits}~\citep{stahlberg2020seq2edits} & Yes & Yes & 58.6 & - \\
\textit{GECToR}(\textit{RoBERTa})~\citep{omelianchuk2020gector} & Yes & Yes &  64.0 & \bf \boldmath 12.4$\times$ \\
\textit{GECToR}(\textit{XLNet})~\citep{omelianchuk2020gector} & Yes & Yes & 65.3 &  - \\ \hline
12+2 BART-Init (beam=1) & Yes & Yes & \bf 66.4 &  2.3$\times$ \\ 
\bf 12+2 BART-Init (IAD) & Yes & Yes & \bf 66.4 &  \bf 9.6$\times$ \\ 
\hline
\end{tabular}
\vspace{0.15cm}
\caption{The performance and online inference efficiency evaluation of efficient GEC models in CoNLL-14. For the models with $\star$, their performance and speedup numbers are from \cite{chen2020improving} who evaluate the online efficiency in the same runtime setting (e.g., GPU and runtime libraries) with ours. The underlines indicate the speedup numbers of the models are evaluated with Tensorflow based on their released codes, which are not strictly comparable here. Note that for \textit{GECToR}, we re-implement its inference process of \textit{GECToR} (RoBERTa) using fairseq for testing its speedup in our setting. - means the speedup cannot be tested in our runtime environment because the model has not been released or not implemented in fairseq.} \label{tab:finalenglish}
\end{table}

\begin{table}[t]
\centering
\begin{tabular}{l|cc|ccc}
\hline
\multirow{2}{*}{\textbf{Model}} &  \multicolumn{2}{c|}{\textbf{NLPCC-18} (big)} & \multicolumn{3}{c}{\textbf{Wikilarge} (base)} \\ 
 & $F_{0.5}$ & Speedup & SARI & BLEU & Speedup \\ \hline
\it Transformer (beam=5) &  29.6 & 1.0$\times$ & 36.9 & 91.9 & 1.0$\times$ \\
\it Transformer (beam=1) &  29.4 & 1.1$\times$ & 36.1 & 90.7 & 1.1$\times$ \\\hline
\it Transformer (IAD) & 29.4 & \bf 8.5$\times$ & 36.1 & 90.7 & \bf 8.8$\times$ \\ \hline
\end{tabular}
\vspace{0.1cm}
\caption{The results of IAD for Chinese GEC (NLPCC-18) and English Text Simplification (Wikilarge) when it is applied to a naive Transformer baseline: For Chinese GEC, we use Transformer-big, while for English Text Simplification, we use Transformer-base as the baseline.}\label{tab:chinese}
\end{table}

Table~\ref{tab:finalenglish} shows the performance and efficiency of IAD applied to a BART-initialized \citep{Lewis:2020} Transformer with a 12-layer encoder and 2-layer decoder that is pretrained with synthetic data and fine-tuned in multiple stages \citep{stahlberg2020seq2edits}, and compares with state-of-the-art efficient GEC models that are all faster than the Transformer-big baseline in CoNLL-14 test set. IAD significantly speeds up the 12+2 BART-initialized single model\footnote{The same model checkpoint also achieves the state-of-the-art result -- 72.9 $F_{0.5}$ with a $9.3\times$ speedup in the BEA-19 test set.} to achieve a $9.6\times$ speedup over the Transformer-big baseline without hurting its state-of-the-art result. 

Unlike \textit{GECToR} and \textit{PIE} that are difficult to adapt to other languages and tasks despite their competitive speed because they are specially designed for English GEC with many manually designed language-specific operations like the transformation of verb forms (e.g.,
VBD$\to$VBZ) and prepositions (e.g., in$\to$at), IAD is data-driven without depending on language-specific features, and thus can be easily adapted to other languages (e.g., Chinese) and tasks (e.g., text simplification) whose inputs and outputs are similar, both achieving an over $8\times$ speedup with identical results as autoregressive (greedy) decoding, as shown in Table \ref{tab:chinese}.

\subsection{Evaluations for Generalized Aggressive Decoding}\label{subsec:gad_exp}
\subsubsection{Experimental Settings}
\label{sec:settings}
\paragraph{Datasets and Evaluation}
We mainly evaluate Generalized Aggressive Decoding (GAD) on the most recognized machine translation benchmark: WMT14\footnote{\url{https://www.statmt.org/wmt14}} English$\to$German translation which contains 4.5M translation pairs for training. Following prior work~\citep{ott-etal-2018-scaling}, we adopt \textit{newstest-13} as our validation set for finding the best hyperparameters and model checkpoints, and test on \textit{newstest-14}. We use 32K Byte Pair Encoding (BPE)~\citep{sennrich2016neural} subwords as the joint source-target dictionary. We use BLEU~\citep{Papineni:2002} to evaluate the translation quality. For inference efficiency, we use both the number of decoding iterations and the speedup over the beam search baseline. Specifically, we test the inference speed with fairseq implementation using Pytorch 1.10 with CUDA 11 on 1 Nvidia P100 GPU.

In addition to WMT14 English$\to$ German, we also test GAD on WMT14 German$\to$English and WMT16 English$\leftrightarrow$ Romanian translation benchmarks, following previous NAR work in machine translation.

\paragraph{Model Configuration}
We mainly study the most commonly used base-size Transformer~\citep{vaswani2017attention} architecture for MT. The Transformer-base has a 6-layer encoder and a 6-layer decoder. Its embedding/FFN dimension/\#heads are 512/2,048/8. We use the model architecture for both the drafter (NAR) and verifier (AR). We apply sequence-level knowledge distillation as discussed in Section \ref{subsec:nat} and \ref{sec:AT-verifier} for the drafter and verifier using a shared teacher: Following recent NAR work~\citep{Ghazvininejad:2019,Saharia:2020,Savinov:2022}, we use the Transformer-big as the teacher for WMT14 EN$\leftrightarrow$DE, and use Transformer-base for WMT16 EN$\leftrightarrow$RO, which all train with the raw training set and generate the distilled training set with beam search (\textit{beam} $=5$). We include model training details in Appendix \ref{subsec:gad_hyperparameters}.

\begin{table}[t]
\small
\centering
\scalebox{0.87}{
\begin{tabular}{clcccccc}
\toprule
\multicolumn{2}{c}{\multirow{2}{*}{\textbf{Models}}} & \multirow{2}{*}{\textbf{Iter.}} &\multirow{2}{*}{\textbf{Speed}} & \multicolumn{2}{c}{\textbf{WMT14}} &\multicolumn{2}{c}{\textbf{WMT16}}\\
\multicolumn{2}{c}{} & & &\textbf{EN-DE} &\textbf{DE-EN} &\textbf{EN-RO} &\textbf{RO-EN}\\\hline
AR baseline & Transformer-base (w/o Seq-KD)$^\dag$ &N &$1.0\times$ &27.38 &31.78 &34.16 &34.46 \\ \hline
\multirow{9}{*}{Fully NAR}
&NAT w/ Fertility \citep{gu:2018} &1 &15.6$\times$ &17.69 &21.47 &27.29 &29.06 \\
&CTC \citep{Libovicky:2018} &1 &/ &17.68 &19.80 &19.93 &24.71 \\
&Bag-of-ngrams \citep{Shao:2020} &1 &10.8$\times$ &20.90 &24.61 &28.31 &29.29 \\
&AXE \citep{Ghazvininejad:2020a}$^*$ &1 &/ &23.53 &27.90 &30.75 &31.54 \\ 
&GLAT \citep{Qian:2020} &1 &15.3$\times$ &25.21 &29.84 &31.19 &32.04 \\ 
&OaXE \citep{Du:2021}$^*$ &1 &/ &26.10 &30.20 &32.40 &33.30 \\ 
&AligNART \citep{Song:2021} &1 &13.4$\times$ &26.40 &30.40 &32.50 &33.10 \\ 
&DSLP \citep{Huang:2021} &1 &14.8$\times$ &27.02 &\textbf{31.61} &\textbf{34.17} &\textbf{34.60} \\ 
&F-VAE \citep{Gu:2021} &1&16.5$\times$  &\textbf{27.49} &31.10 &33.79 &33.87 \\ 
\hline 
\multirow{10}{*}{Iterative NAR}
&iNAT \citep{Lee:2018} &10 &/ &21.61 &25.48 &29.32 &30.19 \\ 
&CMLM \citep{Ghazvininejad:2019}$^*$ &10 &1.7$\times$ &27.03 &30.53 &33.08 &33.31 \\ 
&LevT \citep{Gu:2019} &2.1 &4.0$\times$ &27.27 &/ &/ &33.26 \\ 
&SMART \citep{Ghazvininejad:2020b}$^*$ &10 &1.7$\times$ &27.65 &31.27 &/ &/ \\ 
&DisCo \citep{Kasai:2020}$^*$ &4.8 &3.5$\times$ &27.34 &31.31 &33.22 &33.25 \\ 
&Imputer \citep{Saharia:2020}$^*$ &8 &3.9$\times^\S$ &28.20 &31.80 &34.40 &34.10 \\ 
&Multi-Task NAT \citep{Hao:2021}$^*$ &10 &1.7$\times$ &27.98 &31.27 &33.80 &33.60 \\ 
&RewriteNAT \citep{Geng:2021}$^*$ &2.7 &3.1$\times$ &27.83 &31.52 &33.63 &34.09  \\ 
&SUNDAE \citep{Savinov:2022}$^*$ &16 &1.4$\times^\S$ &28.46 &\color{red}{32.30} &/ &/ \\ 
&CMLMC \citep{Huang:2022}$^*$ &10 &/ &28.37 &31.41 &34.57 &34.13 \\ 
\hline 
\multirow{7}{*}{Ours}
& Teacher &N &/ &29.31 &32.11 &34.72 &34.49 \\ \cline{2-8}
&NAR drafter ($k=25$) &1.6 &$14.3\times$ &26.48 &30.23 &32.08 &32.21 \\  \cline{2-8}
&AR verifier (\textit{beam }$=5$) &N &$1.0\times$ &28.89 &32.53 &34.96 &34.86 \\ 
&AR verifier (\textit{beam }$=1$) &N &$1.1\times$ &28.73 &32.18 &34.83 &34.65 \\ \cline{2-8}
&GAD ($k=25$) &4.9 &3.0$\times$ &\color{red}{28.73} &\color{red}{32.18} &\color{red}{34.83} &\color{red}{34.65} \\ 
&GAD++ ($k=25$, \textit{high-quality}) &4.0 &3.6$\times$ &\color{red}{\textbf{28.89}} &\color{red}{\textbf{32.56}} &\color{red}{\textbf{35.32}} &\color{red}{\textbf{34.98}} \\ 
&GAD++ ($k=25$, \textit{high-efficiency}) &3.1 &\textbf{4.5}$\times$ & \color{red}{28.73} & \color{red}{32.19} & \color{red}{34.92} & \color{red}{34.80} \\
\bottomrule
\end{tabular}
}
\vspace{0.1cm}
\caption{Results of GAD on WMT14 EN$\leftrightarrow$DE and WMT16 EN$\leftrightarrow$RO benchmarks. For inference efficiency, we report the averaged decoding iteration (denoted as \textit{Iter.}) and speedup on WMT14 EN$\rightarrow$DE. The results that truly match or outperform autoregressive decoding (i.e., AR verifier with beam=1) are \textcolor{red}{highlighted in red}. $^\dag$AR baselines' results are from \citet{Kasai:2020}. $^\S$ indicates the speedup results reported in the original papers are obtained by the comparison with greedy decoding.}
\label{tab:exp}
\end{table}

\subsubsection{Main Results}
The translation quality and speedup results in the WMT'14 English-German translation are presented in Table \ref{tab:exp}. Unlike previous NAR approaches that are inferior to AR with Seq-KD (i.e., our AR verifier), GAD introduces an around $3\times$ speedup with the exactly same translation quality as (autoregressive) greedy decoding by our AR verifier, truly achieving lossless acceleration. GAD++ further improves the results by loosening the strict top-1 matching criterion: slightly loosening (i.e., GAD++ high-quality) allow us to achieves better translation than greedy decoding with a higher speedup (3.0$\times$ $\rightarrow$ 3.6$\times$), while a little more aggressively loosening (i.e., GAD++ high-efficiency) further accelerates inference (3.6$\times$ $\rightarrow$ 4.5$\times$) owing to the acceptance of more tokens despite a marginal loss of translation quality.

By looking into our results, it is observed that our NAR drafter's translation quality is better than the majority of fully NAR work but inferior to most iterative NAR approaches. Compared with the NAR models including complicated mechanisms such as length prediction, length beam, reranking, and CTC that slow down the efficiency per iteration, our NAR drafter is simple and straightforward. As a result, its decoding efficiency per iteration is much higher, leading to a comparable speedup to fully NAR despite taking 1.6 decoding iterations on average. The acceptable translation quality and high efficiency of our NAR drafter significantly help accelerate autoregressive decoding, playing a critical role in the lossless acceleration of GAD.

The results for other language pairs are similar to WMT14 EN-DE. We include the details in Table \ref{tab:exp-details} in Appendix \ref{subsec:gad-exp-details}.

\subsubsection{Analysis of Hyperparameter}
\label{sec:tuned-hyperparameter}

\paragraph{Block Size $k$}
We conduct experiments with various block sizes on our development set and show the results in Table \ref{tab:block-size}. As the block size $k$ increases, the number of mean accepted tokens, which highly correlates with speedup and the number of decoding iterations, first increases and reaches a peak when $k=25$. Further increasing $k$ has an adverse effect, because it will become very hard for the model to learn to translate too many tokens simultaneously given the limited model capacity, leading to a drop in both efficiency and quality.

\begin{table}[t]
\small
\centering
\begin{tabular}{l|cccc}
\toprule
\textbf{Models} &$\boldsymbol{k}$ &\textbf{Tok.} &\textbf{BLEU} &\textbf{Speed} \\ \hline
AR (\textit{beam} $=5$) &/ &1.00 &26.72 &1.00$\times$ \\ \hline
\multirow{5}{*}{GAD++}
&10 &5.97 &26.68 &3.04$\times$ \\ 
&15 &6.74 &\textbf{26.94} &3.47$\times$ \\ 
&20 &7.24 &26.75 &3.55$\times$ \\ 
&25 &\textbf{7.56} &26.92 &\textbf{3.79}$\times$ \\ 
&30 &7.44 &26.75 &3.63$\times$\\ 
\bottomrule
\end{tabular}
\vspace{0.2cm}
\caption{The mean accepted tokens (\textit{Tok.}), the translation quality (\textit{BLEU}), and the efficiency (\textit{Speed}) when decoding with a various number of block size $k$ on the development set. The results are obtained with GAD++ (\textit{top-3}, $\tau=1.0$).}
\label{tab:block-size}
\end{table}

\begin{table}[t]
\small
\centering
\begin{tabular}{l|cccc}
\toprule
\textbf{Models} &$\boldsymbol{\tau}$ &\textbf{Top-3} ($\beta=3$) &\textbf{Top-5} ($\beta=5$) \\ \hline
\multirow{5}{*}{GAD++}
&1 &7.56/\textbf{27.02} &7.58/27.02\\ 
&2 &8.64/26.92 &8.77/26.92\\ 
&3 &9.46/26.84 &9.72/26.84\\ 
&4 &10.04/26.78 &10.50/26.74\\ 
&5 &10.38/26.70 &\textbf{10.99}/26.64\\
\bottomrule
\end{tabular}
\vspace{0.15cm}
\caption{Performances on the development set with different hyperparameters in GAD++. Each cell lists the mean accepted tokens and BLEU score. The results are obtained with GAD++ ($k=25$). The BLEU score of greedy decoding of the AR verifier is 26.62.}
\label{tab:GAD++}
\end{table}

\paragraph{Top-$\beta$ and Tolerance $\tau$ in GAD++}
We study the effects of hyperparameters in GAD++: top-$\beta$ and tolerance $\tau$, and show the results on the development set in Table \ref{tab:GAD++}. Moderately increasing $\tau$ and $\beta$ not only leads to an increase of mean accepted tokens since AR verification becomes less strict but also improves the translation quality over greedy decoding. However, the translation quality will decrease if the constraints are over loosened: the BLEU score will degrade from the peak of 27.02 to 26.64 when decoding with \textit{top-5} selection (i.e., $\beta=5$) and $\tau=5.0$. Based on the results in the development set, we conservatively select $\beta=3, \tau=1.0$ for the \textit{high-quality} GAD++, and use $\beta=5, \tau=3.0$ as the \textit{high-efficiency} GAD++ to pursue the higher speedup without substantial loss of translation quality for WMT14 English$\to$German translation as in Table \ref{tab:exp}.

\subsubsection{Analysis of Model Size}

In addition to the base-size models, we also study larger models to test the effectiveness of GAD. We here use Transformer-big~\citep{vaswani2017attention} as our model architecture for both the NAR drafter and AR verifier in GAD/GAD++\footnote{The hyperparameters (e.g., block size $k$, Top-$\beta$, tolerance $\tau$) in GAD/GAD++ (big) are re-tuned on the development set, which may be different from those in the base-size models.}, and compare it with the conventional Transformer-big baseline as well as Blockwise Decoding~\citep{Stern:2018} -- a state-of-the-art efficient Transformer-big variant by introducing additional $k-1$ heads on top of the Transformer decoder to generate next $k$ tokens as a block and verifies, which works in a similar way to GAD. According to Table \ref{tab:exp2}, our GAD/GAD++ substantially outperforms the baseline and Blockwise Decoding, introducing a $3\times$$\sim$$5\times$ speedup over the Transformer-big beam search. Compared with Blockwise Decoding that only uses lightweight heads to generate the next few tokens in parallel, our independent NAR drafter in GAD is much more powerful to generate more drafted tokens that can be accepted\footnote{Our vanilla GAD (base) accepts an average number of 6.13 tokens at each decoding iteration on the development set, and GAD++ increases the number to 10.99 tokens with comparable quality, while Blockwise Decoding (base) is reported to accept only 1.90 tokens on average at each decoding iteration with no quality loss.}, 
leading to a significantly higher speedup, despite the introduction of more parameters that only account for negligible additional memory cost (see Table \ref{tab:memory} in Appendix \ref{sec:gpu_memory}).

\begin{table}[t]
\small
\centering
\begin{tabular}{cl|ccc}
\toprule
\multicolumn{2}{c|}{\textbf{Models}} &\textbf{Iteration} &\textbf{BLEU} &\textbf{Speed} \\ \hline
Teacher &Transformer-big (\textit{beam} $=5$) &N &29.31 &1.0$\times$ \\\hline 
\multirow{2}{*}{\begin{tabular}[c]{@{}c@{}}Blockwise \\\cite{Stern:2018}\end{tabular}}
&Blockwise decoding ($k=2$)&/ &28.95 &1.7$\times^\dag$ \\
&Blockwise decoding ($k=10$)&/ &27.40 &3.0$\times^\dag$ \\
\hline 
\multirow{6}{*}{Ours}
&NAR drafter ($k=30$) &1.4 &27.35 &$15.0\times$ \\  \cline{2-5}
&AR verifier (\textit{beam} $=5$) &N &29.25 &$1.0\times$ \\ 
&AR verifier (\textit{beam} $=1$) &N &29.18 &$1.1\times$ \\ \cline{2-5}
&GAD ($k=30$) &4.8 &29.18 &3.0$\times$ \\
&GAD++ ($k=30$, \textit{top-3}, $\tau=1.0$) &4.2 &\textbf{29.32} &3.5$\times$ \\
&GAD++ ($k=30$, \textit{top-5}, $\tau=6.0$) &2.6 &29.15 &\textbf{5.0}$\times$ \\ 
\bottomrule
\end{tabular}
\vspace{0.15cm}
\caption{Results of GAD of the big-size model configuration on WMT14 EN-DE and the comparison to the state-of-the-art Blockwise Decoding \cite{Stern:2018}. $^\dag$ denotes the speedup results reported in original papers obtained by comparison with greedy decoding.}
\label{tab:exp2}
\end{table}

Moreover, we observe that big-size models can use a larger block size ($k=30$) than the base-size models ($k=25$) since larger capacity equips the model with a more powerful ability to learn to decode more tokens well in parallel. To better demonstrate this point, we conduct a comparative study of the effects of the NAR drafter's size given the same block size ($k=30$) in the GAD-base setting. According to Table \ref{tab:stronger-nat}, the big-size NAR drafter largely outperforms the base-size counterpart: it can generate drafted tokens more reliably (i.e., on average more drafted tokens accepted by the AR verifier), resulting in fewer decoding iterations, which indicates that GAD can be further improved if a more powerful NAR drafter is equipped.

\begin{table}[t]
\centering
\begin{tabular}{lcccc}
\toprule
\textbf{Models} & \textbf{Tok.} & \textbf{Iter.} & \textbf{BLEU}\\ \hline
AR-base (greedy) &1 &N &28.73 \\ \hline
GAD-base &5.53 &5.0 &28.73 \\
$\llcorner$ \textit{w/} NAR-big &\textbf{5.90} &\textbf{4.7} &\textbf{28.73}\\ \hline
GAD++-base (\textit{top-3}, $\tau=1.0$) &6.69 &4.1 &28.81 \\
$\llcorner$ \textit{w/} NAR-big &\textbf{7.32} &\textbf{3.8} &\textbf{29.12}\\ \hline
GAD++-base (\textit{top-5}, $\tau=3.0$) &8.71 &3.2 &28.58 \\
$\llcorner$ \textit{w/} NAR-big  &\textbf{9.73} &\textbf{2.8} &\textbf{28.98} \\
\bottomrule
\end{tabular}
\vspace{0.1cm}
\caption{A comparative study on WMT14 EN-DE test set by replacing NAR-base in GAD-base with the stronger NAR-big. The results are obtained with GAD ($k=30$).}
\label{tab:stronger-nat}
\end{table}

\subsubsection{Analysis of Other Seq2seq Tasks}

We test GAD's effectiveness in one of the most representative seq2seq task -- Abstractive Summarization. We employ the distilled training data of CNN Daily Mail \citep{Hermann:2015} from BART \citep{Lewis:2020} to train the NAR drafter and AR verifier that are both based on BART-base, and test on the CNN Daily Mail test split following previous work.

According to Table \ref{tab:sum}, vanilla GAD consistently achieves the exactly same result as AR verifier (\textit{beam} $=1$), which is, to the best of our knowledge, the first work that achieves such a $3\times$ lossless speedup for Abstractive Summarization. GAD++ further accelerates inference but does not show any quality improvement as observed in MT experiments because of the larger performance gap between the NAR drafter and the AR verifier.

\begin{table}[t]
\centering
\small
\begin{tabular}{cl|ccccc}
\toprule
\multicolumn{2}{c|}{\textbf{Models}} &\textbf{Iteration} &\textbf{R-1} &\textbf{R-2} &\textbf{R-L} &\textbf{Speed} \\ \hline
Teacher &BART \citep{Lewis:2020} &N &44.16 &21.28 &40.90 &/ \\\hline 
AR baseline &BART-base (\textit{beam} $=5$) w/o Seq-KD &N &42.84 &20.08 &39.51 &1.0$\times$ \\\hline 
\multirow{6}{*}{Ours}
&NAR drafter ($k=25$) &3.3 &37.10 &14.87 &33.47 &15.0$\times$ \\  \cline{2-7}
&AR verifier (\textit{beam} $=5$) &N &42.55 &19.83 &39.31 &1.0$\times$ \\ 
&AR verifier (\textit{beam} $=1$) &N &43.00 &20.28 &39.96 &1.1$\times$ \\ \cline{2-7}
&GAD ($k=25$) &14.0 &\textbf{43.00} &\textbf{20.28} &\textbf{39.96} &3.0$\times$ \\
&GAD++ ($k=25$, \textit{top-3}, $\tau=1.0$) &10.8 &42.95 &20.24 &39.73 &3.7$\times$ \\
&GAD++ ($k=25$, \textit{top-5}, $\tau=3.0$) &7.9 &42.02 &19.42 &38.66 &\textbf{4.8}$\times$ \\ 
\bottomrule
\end{tabular}
\vspace{0.1cm}
\caption{Results of GAD/GAD++ on CNN-DM for Abstractive Summarization.}
\label{tab:sum}
\end{table}

\subsection{Discussion}\label{subsec:discuss}
\begin{figure}[t]
\centering
\includegraphics[width=0.8\textwidth]{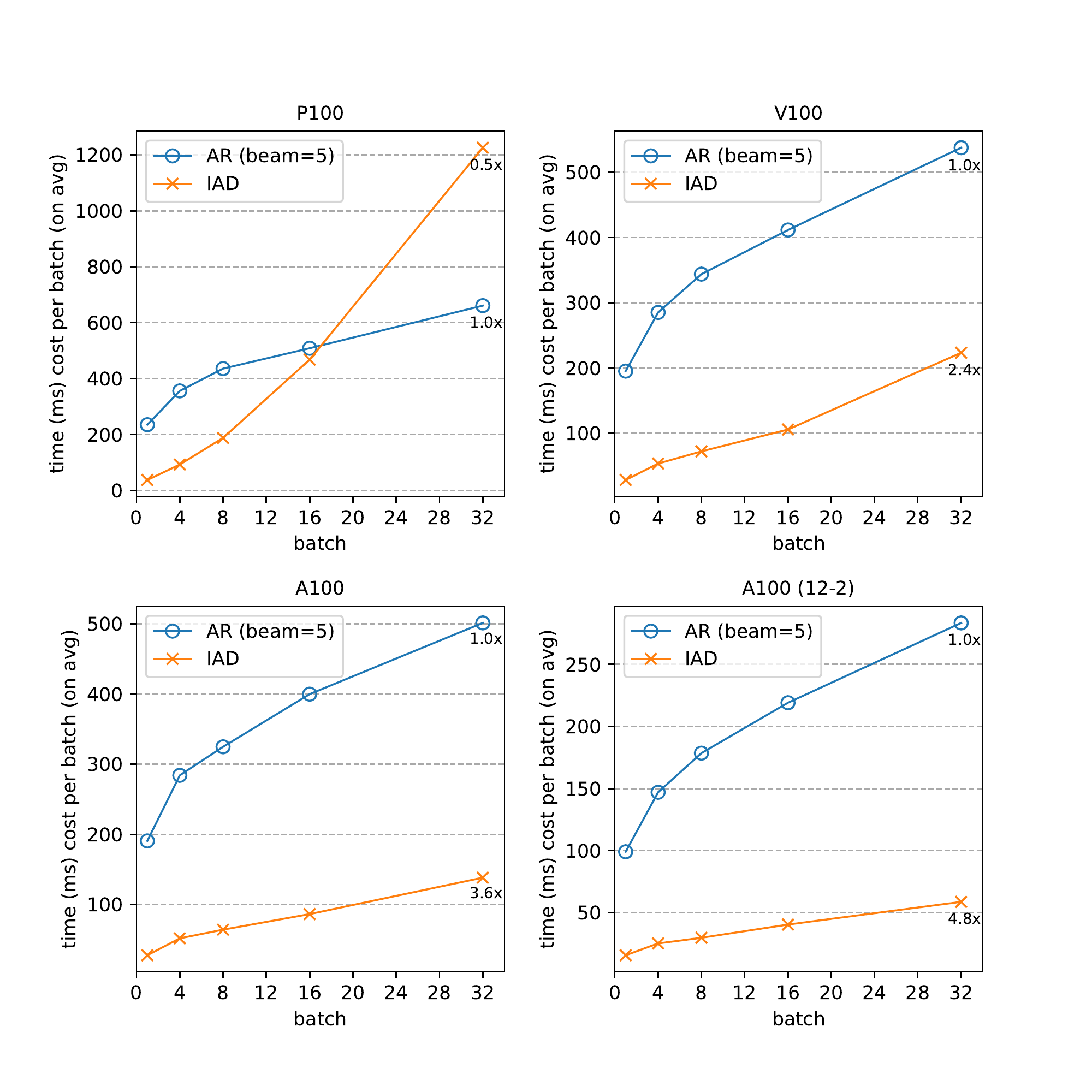}
\vspace{-0.3cm}
\caption{Inference latency of IAD for the Transformer-big (6-layer encoder and 6-layer decoder) and the 12+2 BART-initialized Transformer with batch implementation on P100 (fp32), V100 (fp16) and A100 (fp16). The results are obtained on the CoNLL-14 test set. The speedup baseline ($1\times$) is AR decoding with beam search (\textit{beam} $=5$) when batch $=32$.}
\label{fig:batch-iad}
\end{figure}

\begin{figure}[t]
\centering
\includegraphics[width=1.0\textwidth]{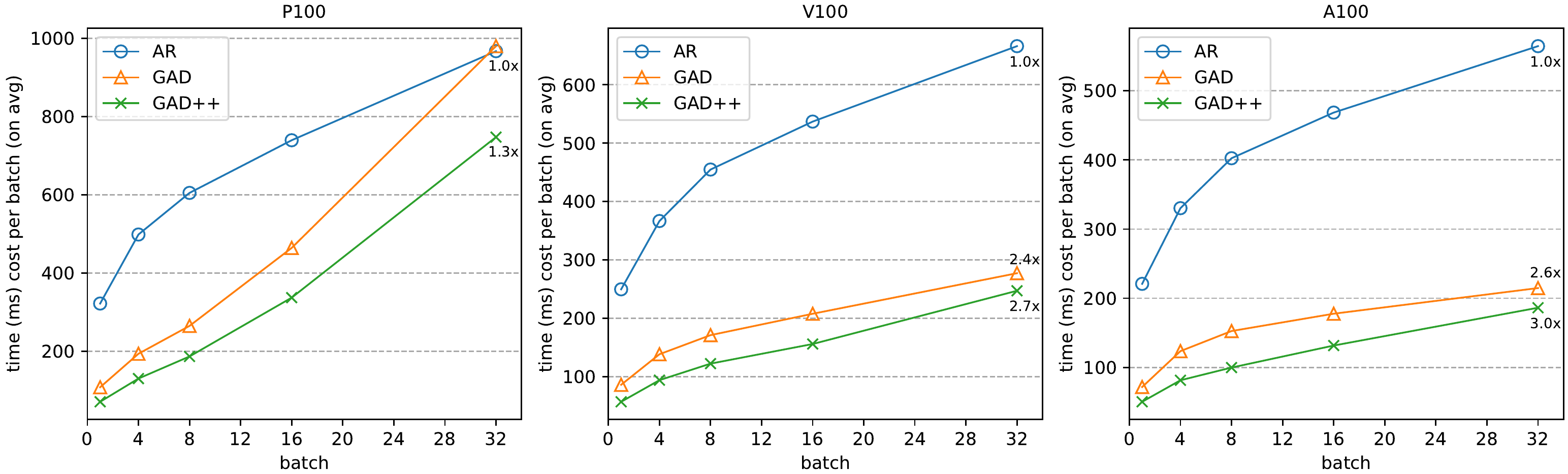}
\caption{Inference latency of GAD/GAD++ with batch implementation on P100 (fp32), V100 (fp16) and A100 (fp16). The results are obtained on the WMT14 EN-DE test set with GAD/GAD++-Base ($k=25$, \textit{top-5}, $\tau=3.0$). The speedup baseline ($1\times$) is AR (\textit{beam} $=5$) when batch $=32$.}
\label{fig:batch-gad}
\end{figure}

Extensive experiments in multiple tasks show that Aggressive Decoding can significantly speed up seq2seq generation without quality loss: IAD introduces a $7\times$$\sim$$9\times$ speedup for the seq2seq tasks characterized by highly similar inputs and outputs, and GAD/GAD++ can have a $3\times$$\sim$$5\times$ speedup which is slower than IAD because of the additional time cost of the NAR drafter in typical Machine Translation and Abstractive Summarization benchmarks. The state-of-the-art lossless speedup is attributed to the substantially improved computational parallelism that allows better utilization of (hardware) computing resources. We believe Aggressive Decoding is promising and can even benefit more from evolving processor hardware that will become increasingly powerful and better at parallel computing, as shown in the speedup comparison of batch inference\footnote{As the batch size increases, the inference latency per batch will become higher. Therefore, it is impractical to use a large batch size during inference, as \citet{rajbhandari2022deepspeed} points out. Therefore, we use the batch size of up to 32 in our experiments. Please note that our batch inference is currently a naive and under-optimized implementation without using cached attention key/value to save the computation of previous decoded tokens for verification, which means its result is probably much underestimated and should be more efficient with an optimized implementation.} using P100, V100 and A100 in Figure \ref{fig:batch-iad} and \ref{fig:batch-gad}.

As a preliminary study, Aggressive Decoding, especially GAD, is far from perfect and has much room to improve. For example, according to the experimental results above, we know GAD's translation quality mainly depends on the AR verifier and its efficiency relies on the NAR drafter (whose capability matters how many drafted tokens can be accepted). We believe more powerful NAR/AR models (than the simple and naive ones used in this paper) must benefit GAD to achieve better results. 
Also, according to Table \ref{tab:profling} showing time cost by modules in GAD++, our naive implementation costs a total of approximately 16\% of overall time to (sequentially) encode the input for AR and NAR. Obviously, this part can be optimized by performing AR and NAR encoding in parallel because they are independent. Moreover, the NAR decoder costs more than the AR decoder because it employs bi-directional attention and cannot save the computation for the already decoded tokens as AR, which we believe can be improved in the future with a better non-autoregressive decoding mechanism designed for GAD.

\section{Related Work}
\paragraph{Non-autoregressive Decoding} To address the inefficiency of autoregressive decoding for seq2seq generation, \citet{gu:2018} first proposed non-autoregressive decoding for Machine Translation, which decodes the output sentence in one single iteration despite translation quality loss. Recent work mainly focused on improving the quality while maintaining competitive speedups, including applying various training objectives \citep{Ghazvininejad:2020a, Saharia:2020, Du:2021, Huang:2021}, modeling dependencies between target tokens \citep{Ghazvininejad:2019, Qian:2020, Song:2021, Gu:2021} and refining the translation outputs with multi-pass iterations \citep{Ghazvininejad:2020b, Kasai:2020, Hao:2021, Geng:2021, Savinov:2022, Huang:2022}. However, due to the inherent conditional independence assumption, non-autoregressive decoding's quality is generally less reliable than the autoregressive counterpart.

\paragraph{Semi-autoregressive Decoding} There are also some attempts trying to combine autoregressive and non-autoregressive decoding: \citet{Wang:2018} proposed to utilize non-autoregressive decoding locally while keeping the autoregressive property globally; on the contrary, \citet{Ran:2020} introduced a local-autoregressive model which retained the non-autoregressive property globally. Similar ideas have been also proposed for GEC: For example, \citet{chen2020improving} proposed to use a sequence tagger to identify the grammatical errors' spans and then use autoregressive decoding to edit them to make the sentence grammatically correct, which is an early form of Input-guided Aggressive Decoding. The most similar work to Generalized Aggressive Decoding is Blockwise Decoding \citep{Stern:2018} that proposed to additionally insert $k-1$ non-autoregressive heads on top of the Transformer decoder to generate $k$ positions in parallel and use the original autoregressive head to verify these outputs. However, its underinvestment in the non-autoregressive modeling seriously limits its performance, resulting in a much lower efficiency than our approach.

\paragraph{Cascade Inference} In general, cascade inference includes model cascade~\citep{Huang:2018, Streeter:2018,wang:2020} that sequentially applies a series of light-weighted models to handle various instances based on their difficulties, and dynamic early exiting~\citep{Xin:2020, Liu:2020, Zhou:2020, Sun:2021} that introduces internal classifiers within deep neural networks to allow early exiting if the prediction is confident enough. Both of them are for efficient inference. Broadly, Aggressive Decoding can be also considered as a special case of cascade inference where aggressively generating drafted tokens is the first round inference and the following verification is analog to the second round responsible for controlling the quality, which should be the very early exploration on seq2seq generation in this direction, to the best of our knowledge.

\section{Conclusion}
We study lossless acceleration for seq2seq generation and propose a novel decoding paradigm -- Aggressive Decoding which replaces the conventional step-by-step decoding with aggressive decoding and verification. Our proposed two Aggressive Decoding variants -- Input-guided Aggressive Decoding (IAD) and Generalized Aggressive Decoding (GAD) -- can both achieve lossless generation quality in a significant speedup in the rewriting (e.g., GEC) and general seq2seq tasks (e.g., MT) respectively.

Despite the state-of-the-art lossless speedup results, Aggressive Decoding still has great potential with much room to improve, as discussed in Section \ref{subsec:discuss}. We hope that our preliminary study could draw more attention to this promising decoding paradigm that may potentially evolve into a \textit{de facto} standard for efficient and lossless seq2seq generation in the near future.

\bibliography{ad}
\bibliographystyle{iclr2022_conference}

\appendix

\begin{table}[!t]
\centering
\small
\begin{tabular} {lr} 
\hline
Configurations	         &	Values
\\
\hline 
\multicolumn{2}{c}{\bf Train From Scratch} \\
\hline
Model Architecture 	    & Transformer-big   \\
Number of epochs		& 60				\\
Devices                 & 4 Nvidia V100 GPU      \\
Max tokens per GPU		& 5120					\\
Update Frequency        & 4                     \\
Optimizer 				& Adam 					\\
						& ($\beta_1$=0.9, $\beta_2$=0.98, $\epsilon$=$1\times10^{-8}$)	\\
Learning rate 			& [$3\times10^{-4}$ , $5\times10^{-4}$] \\
Learning rate scheduler & inverse sqrt \\ 
Warmup                  & 4000 \\
Weight decay            & 0.0 \\
Loss Function 			& label smoothed cross entropy \\
						& (label-smoothing=0.1) \\
Dropout 				& [0.3, 0.4, 0.5] \\
\hline
\multicolumn{2}{c}{\bf Pretrain} \\
\hline
Number of epochs 		& 10					\\
Update Frequency        & 16                    \\
Learning rate 			& $3\times10^{-4}$			\\
Warmup                  & 8000 \\
Dropout					& 0.3	\\
\hline
\multicolumn{2}{c}{\bf Fine-tune} \\
\hline
Number of epochs 		& 60					\\
Update Frequency        & 4                     \\
Learning rate 			& $3\times10^{-4}$			\\
Warmup                  & 4000 \\
Dropout					& 0.3	\\
\hline
\end{tabular}
\vspace{0.1cm}
\caption{Hyperparameters values of training from scratch, pretraining and fine-tuning for GEC. \label{tab:gec_param}}
\end{table}

\section{Details of Evaluation of Input-guided Aggressive Decoding}\label{sec:app_iad}

\subsection{Hyperparameters}

Hyperparameters of training the Transformer for English GEC are listed in Table \ref{tab:gec_param}. The hyperparameters for Chinese GEC are the same as those of training from scratch. 

\subsection{IAD on CPU}


\begin{table}[h]
\centering
\begin{tabular}{c|c|c|c}
\hline
\multirow{2}{*}{\textbf{\begin{tabular}[c]{@{}c@{}}Model\\ (Enc+Dec)\end{tabular}}} &
\multirow{2}{*}{\textbf{Thread}} &
\textbf{Beam=5} & \textbf{Aggressive} \\
 & & Speedup & Speedup \\
 \hline
6+6 & 8 & 1$\times$ & 6.5$\times$ \\
\hline
6+6 & 2 & 1$\times$ & 6.1$\times$ \\
\hline
\end{tabular}
\vspace{0.15cm}
\caption{The efficiency of the 6-layer Transformer in CoNLL-13 on CPU with 8 and 2 threads. \label{tab:cpu} }
\end{table}

Table~\ref{tab:cpu} shows the IAD's speedup for the 6-layer Transformer on an Intel\textsuperscript{\tiny\textregistered} Xeon\textsuperscript{\tiny\textregistered} E5-2690 v4 Processor (2.60GHz) with 8 and 2 threads\footnote{We set OMP\_NUM\_THREADS to 8 or 2.}, respectively. IAD achieves an over $6\times$ online (batch=1) inference speedup over the Transformer-big baseline on CPU. 

\section{Details of Evaluation of Generalized Aggressive Decoding}\label{sec:app_gad}

\subsection{Hyperparameters}
\label{subsec:gad_hyperparameters}
Hyperparameters of training GAD are listed in Table \ref{tab:hyperparameters}. Following \citet{vaswani2017attention} and \citet{ott-etal-2018-scaling}, we also average model parameters from the last 10 checkpoints.

\begin{table}[t]
\centering
\begin{tabular}{lr}
\toprule
\textbf{Hyperparameter} &\textbf{Value} \\\hline
devices &8 Nvidia V100 GPU\\
label smoothing &0.1  \\
\# max tokens &20000  \\
update frequency &4  \\
dropout rate &[0.1, 0.2, 0.3]  \\
Adam lr & $1 \times 10^{-3}$  \\
Adam $\beta_{1}$ &0.9  \\
Adam $\beta_{2}$ &0.99  \\
lr-scheduler &inverse square \\
weight decay &0.00001  \\
clip norm &3.0  \\
\# warmup updates &4000  \\
max updates &100K  \\
max epoch &1000  \\
 \bottomrule
\end{tabular}
\quad
\begin{tabular}{lr}
\toprule
\textbf{Hyperparameter} &\textbf{Value} \\\hline
devices &8 Nvidia V100 GPU\\
label smoothing &0.1  \\
\# max tokens &4096  \\
update frequency &4  \\
dropout rate &[0.1, 0.2, 0.3]  \\
Adam lr & $5 \times 10^{-4}$  \\
Adam $\beta_{1}$ &0.9  \\
Adam $\beta_{2}$ &0.999  \\
Adam $\epsilon$ &$1 \times 10^{-6}$  \\
lr-scheduler &inverse square \\
weight decay &0.01  \\
clip norm &5.0  \\
\# warmup updates &10000  \\
max updates &300K  \\
 \bottomrule
\end{tabular}
\vspace{0.15cm}
\caption{Hyperparameters and settings of the AR verifier (left) and the NAR drafter (right).}
\label{tab:hyperparameters}
\end{table}

\begin{table}[b]
\small
\centering
\scalebox{0.9}{
\begin{tabular}{cl|cccc|cccc}
\toprule
\multicolumn{2}{c|}{\multirow{2}{*}{\textbf{Models}}} & \multicolumn{4}{c|}{\textbf{EN$\rightarrow$X}} & \multicolumn{4}{c}{\textbf{X$\rightarrow$EN}} \\
\multicolumn{2}{c|}{} &\textbf{$\beta,\tau$} &\textbf{Iter.} &\textbf{BLEU} &\textbf{Speed} &\textbf{$\beta,\tau$} &\textbf{Iter.} &\textbf{BLEU} &\textbf{Speed}\\ \hline
AR &AR verifier (\textit{beam} $=1$) &/ &N &28.73 &1.1$\times$  &/ &N &32.18 &1.1$\times$ \\\hline
\multirow{4}{*}{\begin{tabular}[c]{@{}c@{}}WMT14 \\EN-DE\end{tabular}}
&NAR drafter &/ &1.6 &26.48 &14.3$\times$ &/ &1.5 &30.23 &14.0$\times$ \\\cline{2-10}
&GAD ($k=25$) &/ &4.9 &28.73 &3.0$\times$ &/ &4.4 &32.18 &3.0$\times$ \\
&GAD++ ($k=25$, \textit{high-quality}) &3,1.0 &4.0 &28.89 &3.6$\times$ &3,1.0 &3.4 &32.56 &3.9$\times$\\
&GAD++ ($k=25$, \textit{high-efficiency}) &5,3.0 &3.1 &28.73 &4.5$\times$ &5,3.0 &2.5 &32.19 &4.8$\times$  \\ \hline
AR &AR verifier (\textit{beam} $=1$) &/ &N &34.83 &1.1$\times$  &/ &N &34.65 &1.1$\times$ \\\hline
\multirow{4}{*}{\begin{tabular}[c]{@{}c@{}}WMT16 \\EN-RO\end{tabular}}
&NAR drafter &/ &1.6 &32.08 &13.3$\times$ &/ &1.6 &32.21 &13.8$\times$ \\\cline{2-10}
&GAD ($k=25$) &/ &6.4 &34.83 &2.2$\times$ &/ &5.9 &34.65 &2.4$\times$ \\
&GAD++ ($k=25$, \textit{high-quality}) &5,4.0 &4.5 &35.32 &3.1$\times$ &3,2.0 &4.6 &34.98 &2.9$\times$\\
&GAD++ ($k=25$, \textit{high-efficiency}) &5,6.0 &4.0 &34.92 &3.3$\times$ &5,3.0 &4.0 &34.80 &3.3$\times$\\
\bottomrule
\end{tabular}
}
\vspace{0.15cm}
\caption{Details of the main results of GAD/GAD++ on WMT14 EN$\leftrightarrow$DE and WMT16 EN$\leftrightarrow$RO benchmarks. \textit{\textbf{X}} denotes the corresponding language in each benchmark (\textit{German} in WMT14 EN-DE and \textit{Romanian} in WMT16 EN-RO). $k$ is the block size. $\beta$ indicates the Top-$\beta$ selection in GAD++ and $\tau$ is the tolerance hyperparameter. All hyperparameters ($k$, $\beta$ and $\tau$) are tuned on the development set of each benchmark.}
\label{tab:exp-details}
\end{table}

\subsection{Detailed results of GAD/GAD++ on WMT14 EN$\leftrightarrow$DE and WMT16 EN$\leftrightarrow$RO}
\label{subsec:gad-exp-details}
We show the details of our main experimental results in Table \ref{tab:exp-details}. As demonstrated in Section \ref{sec:tuned-hyperparameter}, we tune the hyperparamters ($k$, $\beta$ and $\tau$) of GAD/GAD++ on the development set for each direction of benchmarks. According to Table \ref{tab:exp-details}, GAD/GAD++ performs consistently well on all the language pairs despite slight differences in speedup ratios.

\subsection{Speedup Distribution}
\label{sec:sample_speedup}
\begin{figure}[t]
\centering
\includegraphics[width=0.6\columnwidth]{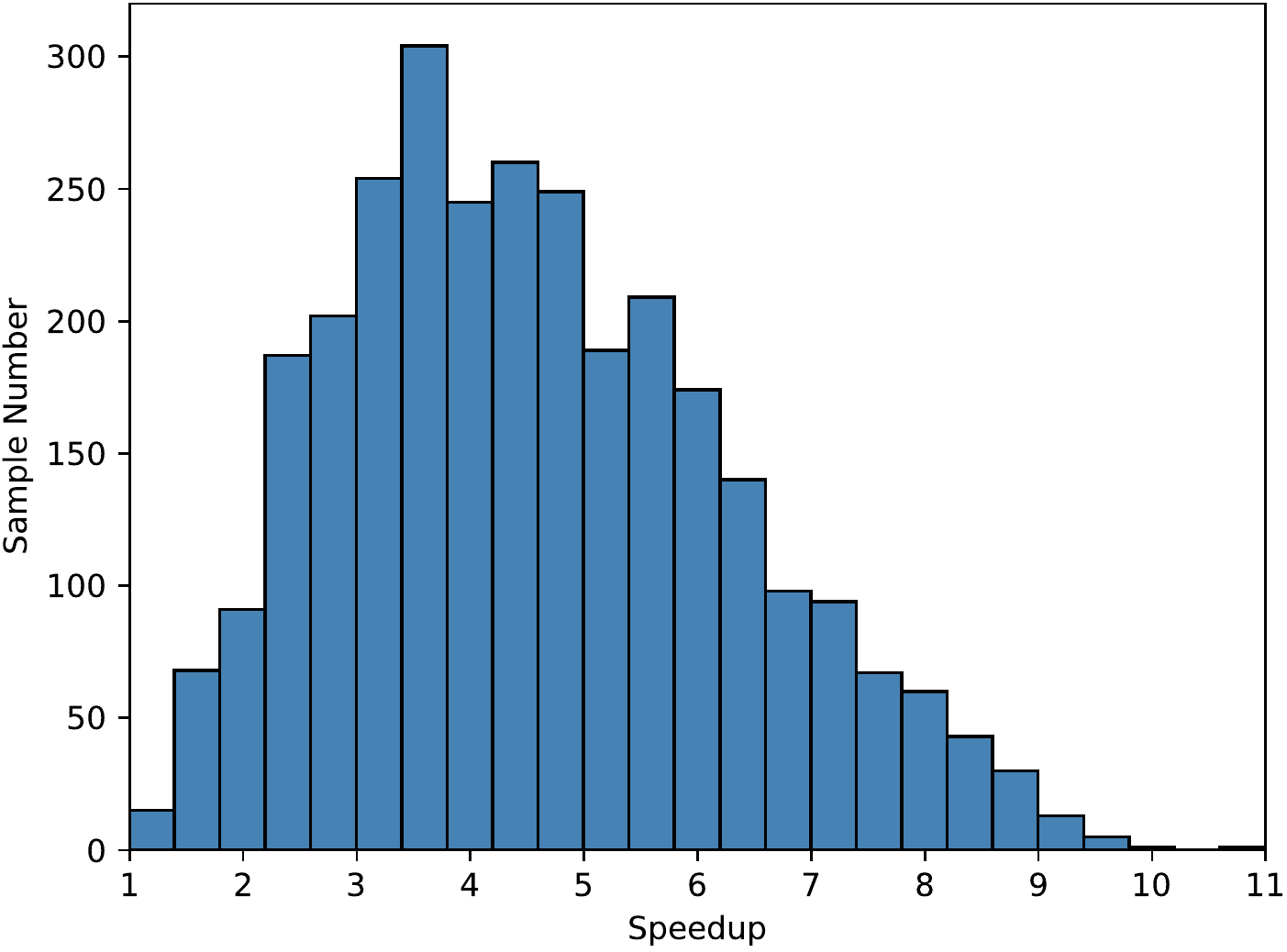}
\caption{Single sentence speedup distribution by GAD++ ($k=25$, \textit{top-5}, $\tau=3.0$) in WMT14 EN-DE test set which has 3,003 sentences in total.}
\label{fig:samplespeed}
\end{figure}

To further understand the acceleration effects of GAD, we present the speedup distribution of a single sentence in the WMT14 EN-DE test set in Figure \ref{fig:samplespeed}, showing that most sentences are translated with a 3$\times$\textasciitilde 6$\times$ speedup compared to the beam search baseline, while some rare cases can even achieve over $10\times$ speedup.

\subsection{Profiling}
\label{sec:profiling}

\begin{table}[h]
\centering
\begin{tabular}{lcc}
\toprule
\textbf{Modules} & \textbf{Latency(ms)} & \textbf{Percent(\%)} \\ \hline
AR Encoder &5.65 &7.70 \\
NAR Encoder &5.73 &7.80 \\
NAR Decoder &31.44 &42.81 \\
AR Decoder &27.33 &37.21 \\
Others &3.31 &4.48 \\\hline
Total &73.46 &100 \\
\bottomrule
\end{tabular}
\vspace{0.15cm}
\caption{Profiling of GAD++ (\textit{base-size}, $k=25$, \textit{top-5}, $\tau=3.0$) on the WMT14 EN-DE test set.}
\label{tab:profling}
\end{table}

We show the inference time cost by modules in GAD++ in Table \ref{tab:profling}. The current naive implementation costs a total of approximately 16\% of overall time to (sequentially) encode the input for AR and NAR, which can be obviously optimized. Also, the NAR decoder costs more than the AR decoder because of the multi-round computation for previously decoded tokens.

\subsection{Memory Analysis}
\label{sec:gpu_memory}

\begin{table}[h]
\centering
\begin{tabular}{lcc}
\toprule
\textbf{Models} & \textbf{Memory Util.} &\textbf{Percent(\%)} \\ \hline
AR (\textit{beam }$=5$) &1237MiB &7.4\\
AR (\textit{beam }$=1$) &1211MiB &7.6\\\hline
GAD &1621MiB &10.0\\ 
GAD++ &1613MiB &9.9\\\hline
Full &16280MiB &100\\
\bottomrule
\end{tabular}
\vspace{0.15cm}
\caption{Comparisons of GPU memory utilization between GAD and AR Baselines. The results are obtained with fp32 computation on a single Nvidia P100 GPU. The hyperparameters of GAD++ are $k=25$, \textit{top-5}, $\tau=3.0$.}
\label{tab:memory}
\end{table}

Table \ref{tab:memory} shows the comparisons of peak GPU memory (footprint) utilization between GAD and AR baselines (during inference). Compared with Transformer-base, GAD-Base only costs about an additional 400MB GPU memory which is negligible for a modern GPU. Among the 400MB additional GPU memory cost, around 250MB is used to store (static) weights of the NAR drafter, and the remaining cost is mainly for storing the final encoder's states of the NAR drafter, which is dynamic and related to the shape (e.g., length) of the input tensor.

\subsection{Case Study}

\begin{table}[t]
\centering
\resizebox{\textwidth}{!}{
\begin{tabular}{|c|l|}
\toprule
\multicolumn{2}{|l|}{\textbf{\textit{Example 1- vanilla GAD}}}  \\\hline
\textsc{Source} & According to the details provided , the tunnel had not yet been put into use .  \\\hline
\textit{D} &Nach den \textcolor{red}{Angaben} Angaben war der Tunnel noch nicht in \\ 
\textit{V} &Nach den \textcolor{red}{vorliegenden} \st{war war der Tunnel noch nicht in Betrieb} \\\hline
\textit{D} &\hl{Nach den vorliegenden} \textcolor{red}{Angaben} war der Tunnel noch nicht in Betrieb genommen\\
\textit{V} &\hl{Nach den vorliegenden} \textcolor{red}{Einzelheiten} \st{war der Tunnel noch nicht in Betrieb genommen worden .}\\\hline
\textit{D} &\hl{Nach den vorliegenden Einzelheiten} war der Tunnel noch nicht in Betrieb genommen \textcolor{red}{{\texttt{[EOS]}}}\\
\textit{V} &\hl{Nach den vorliegenden Einzelheiten} war der Tunnel noch nicht in Betrieb genommen \textcolor{red}{worden}\\\hline
\textit{D} &\hl{Nach den vorliegenden Einzelheiten war der Tunnel noch nicht in Betrieb genommen worden} . {\texttt{[EOS]}}\\
\textit{V} &\hl{Nach den vorliegenden Einzelheiten war der Tunnel noch nicht in Betrieb genommen worden} . {\texttt{[EOS]}}\\\hline
\textsc{Results} &Nach den vorliegenden Einzelheiten war der Tunnel noch nicht in Betrieb genommen worden . \\\hline
\multicolumn{2}{|l|}{\textbf{\textit{Example 2-vanilla GAD}}}  \\\hline
\textsc{Source} & Yesterday , Gut\texttt{@@} acht 's Mayor gave a clear answer to this question .  \\\hline
\textit{D} &Gestern hat \textcolor{red}{der} B{\"u}rger\texttt{@@} meister von Gut\texttt{@@} acht eine klare \\ 
\textit{V} &Gestern hat \textcolor{red}{Gut\texttt{@@}} \st{B{\"u}rger\texttt{@@} meister von Gut\texttt{@@} acht eine klare Antwort} \\ \hline
\textit{D} &\hl{Gestern hat Gut\texttt{@@}} acht\texttt{@@} ts B{\"u}rger\texttt{@@} meister eine klare Antwort auf diese Frage\\
\textit{V} &\hl{Gestern hat Gut\texttt{@@}} \textcolor{red}{ach\texttt{@@}} \st{s B{\"u}rger\texttt{@@} meister eine klare Antwort auf diese Frage gegeben}\\\hline
\textit{D} &\hl{Gestern hat Gut\texttt{@@} ach\texttt{@@}} ts B{\"u}rger\texttt{@@} meister eine klare Antwort auf diese Frage gegeben\\
\textit{V} &\hl{Gestern hat Gut\texttt{@@} ach\texttt{@@}} ts Bürger\texttt{@@} meister eine klare Antwort auf diese Frage gegeben .\\\hline
\textit{D} &\hl{Gestern hat Gut\texttt{@@} ach\texttt{@@} ts B{\"u}rger\texttt{@@} meister eine klare Antwort auf diese Frage gegeben .} {\texttt{[EOS]}}\\
\textit{V} &\hl{Gestern hat Gut\texttt{@@} ach\texttt{@@} ts B{\"u}rger\texttt{@@} meister eine klare Antwort auf diese Frage gegeben .} {\texttt{[EOS]}}\\\hline
\textsc{Results} &Gestern hat Gut\texttt{@@} ach\texttt{@@} ts B{\"u}rger\texttt{@@} meister eine klare Antwort auf diese Frage gegeben . \\\hline
\multicolumn{2}{|l|}{\textbf{\textit{Example 1-GAD++}}}  \\\hline
\textsc{Source} & According to the details provided , the tunnel had not yet been put into use .  \\\hline
\textit{D} &Nach den \textcolor{red}{Angaben} Angaben war der Tunnel noch nicht in \\ 
\textit{V} &Nach den \textcolor{red}{vorliegenden} \st{war war der Tunnel noch nicht in Betrieb} \\\hline
\textit{D} &\hl{Nach den vorliegenden} Angaben war der Tunnel noch nicht in Betrieb genommen \textcolor{red}{{\texttt{[EOS]}}}\\
\textit{V} &\hl{Nach den vorliegenden} \textcolor{blue}{Angaben} war der Tunnel noch nicht in Betrieb genommen \textcolor{red}{worden}\\\hline
\textit{D} &\hl{Nach den vorliegenden Angaben war der Tunnel noch nicht in Betrieb genommen worden} . \texttt{[EOS]}\\
\textit{V} &\hl{Nach den vorliegenden Angaben war der Tunnel noch nicht in Betrieb genommen worden} . \texttt{[EOS]}\\\hline
\textsc{Results} &Nach den vorliegenden Angaben war der Tunnel noch nicht in Betrieb genommen worden . \\\hline
\multicolumn{2}{|l|}{\textbf{\textit{Example 2-GAD++}}}  \\\hline
\textsc{Source} & Yesterday , Gut\texttt{@@} acht 's Mayor gave a clear answer to this question .  \\\hline
\textit{D} &Gestern hat der B{\"u}rger\texttt{@@} meister von Gut\texttt{@@} acht eine klare \\ 
\textit{V} &Gestern hat \textcolor{blue}{der} B{\"u}rger\texttt{@@} meister von Gut\texttt{@@} acht eine klare Antwort \\ \hline
\textit{D} &\hl{Gestern hat der B{\"u}rger\texttt{@@} meister von Gut\texttt{@@} acht eine klare Antwort} auf diese Frage gegeben . \texttt{[EOS]}\\
\textit{V} &\hl{Gestern hat der B{\"u}rger\texttt{@@} meister von Gut\texttt{@@} acht eine klare Antwort} auf diese Frage gegeben . \texttt{[EOS]}\\\hline
\textsc{Results} &Gestern hat der B{\"u}rger\texttt{@@} meister von Gut\texttt{@@} acht eine klare Antwort auf diese Frage gegeben . \\
\bottomrule
\end{tabular}
}
\vspace{0.15cm}
\caption{Examples from the WMT14 English-German translation task. At each iteration, \textit{D} and \textit{V} are the outputs of the drafter and the verifier, respectively. Tokens within \textcolor{red}{red blocks} are the bifurcation positions. The verification pieces after the bifurcation are annotated as strikethrough. The \hl{highlighted parts} are translations of previous iterations. Tokens in \textcolor{blue}{blue blocks} are \textit{top-$\beta$} candidates which meet the GAD++ requirement. The hyperparameters are $k=10$, \textit{top-3}, $\tau=1.0$. `\texttt{@@}' is the BPE token, e.g., Gut\texttt{@@} acht $\rightarrow$ \textit{Gutacht}. The output pieces after the \texttt{[EOS]} token is omitted in the table.} 
\label{tab:case}
\end{table}

Finally, we present examples to illustrate how GAD/GAD++ works in Table \ref{tab:case}.
\end{document}